\documentclass{article}

\usepackage{iclr2025_conference,times}

\usepackage[utf8]{inputenc} % allow utf-8 input
\usepackage[T1]{fontenc}    % use 8-bit T1 fonts

\usepackage{hyperref}       % hyperlinks
\usepackage{url}            % simple URL typesetting
\usepackage{booktabs}       % professional-quality tables
\usepackage{amsfonts}       % blackboard math symbols
\usepackage{nicefrac}       % compact symbols for 1/2, etc.
\usepackage{microtype}      % microtypography

\usepackage{multirow}
\usepackage{booktabs}
\usepackage{makecell}

\usepackage{subcaption}
\usepackage{tcolorbox}
\usepackage{sidecap}
\usepackage{xcolor}  % Add this in the preamble if not already present
\usepackage{amsmath}

\iclrfinalcopy

\usepackage{latexsym}
\usepackage{algorithm}
\usepackage{algpseudocode}
\usepackage{dashrule}
\usepackage{arydshln}
\usepackage{microtype}

\usepackage[inline]{enumitem}

\usepackage{float} % Add this to your preamble

\usepackage{url}            % simple URL typesetting
\usepackage{amsfonts}       % blackboard math symbols
\usepackage{nicefrac}       % compact symbols for 1/2, etc.
\usepackage{microtype}      % microtypography
\usepackage{graphicx}
\usepackage{enumitem}
\usepackage{subcaption}
\usepackage{caption}
\usepackage{rotating}
\usepackage[normalem]{ulem}
\usepackage{amssymb}
\usepackage{colortbl}
\usepackage{linguex}

\usepackage{siunitx} 
\usepackage{pifont}% http://ctan.org/pkg/pifont
\usepackage{tabularx}% added for table design
\usepackage[all]{nowidow}
\definecolor{darkgreen}{rgb}{0.0, 0.42, 0.24}
\definecolor{green}{RGB}{112, 173,71}
\definecolor{blue}{RGB}{68, 114,196}
\definecolor{orange}{RGB}{237, 125,49}
\definecolor{red}{RGB}{202, 54,49}
\definecolor{yellow}{RGB}{222,194, 142}
\usepackage{xspace}

\usepackage{listings}

\definecolor{darkblue}{rgb}{0,0,.5}
\definecolor{darkgreen}{rgb}{0,.5,0}
\definecolor{lightgray}{rgb}{.8,.8,.8}
\definecolor{aliceblue}{rgb}{0.75, 0.75, 1.0}
\definecolor{darkseagreen}{rgb}{0.46, 0.74, 0.46}
\definecolor{alizarin}{rgb}{0.82, 0.1, 0.26}
\definecolor{airforceblue}{rgb}{0.36, 0.54, 0.66}
\definecolor{red_graph}{rgb}{0.98, 0.8, 0.8}
\definecolor{blue_graph}{rgb}{0.8, 0.98, 0.8}
\definecolor{red}{rgb}{0.8, 0.0, 0.0}
\usepackage{wrapfig}
\newcommand{\ignore}[1]{}
\newcommand{\method}{\textsc{Quest}\xspace}
\newcommand{\rlhfmethod}{\textsc{QAlign}\xspace}

\newcommand{\gf}[1]{{\color{red}\textbf{[GF: #1]}}}

\definecolor{darkblue}{rgb}{0, 0, 0.5}
\hypersetup{colorlinks=true, citecolor=darkblue, linkcolor=darkblue, urlcolor=darkblue}

\title{ Sample, Don't Search: \\ Rethinking Test-Time Alignment for Language Models}
\author{%
  {Gonçalo Faria}$^{1}$, {Noah A.~Smith}$^{1,2}$ \\
  ${}^{1}$University of Washington, ${}^{2}$Allen Institute for AI\\
  \texttt{gfaria@cs.washington.edu}
}

\definecolor{purp}{HTML}{791f87}

\definecolor{SFT-BoN}{HTML}{e41a1c}
\definecolor{SFT-MV}{HTML}{377eb8}
\definecolor{SFT-WMV}{HTML}{4daf4a}
\definecolor{DPO-MV}{HTML}{984ea3}
\definecolor{SFT-QUEST}{HTML}{ff7f00}

\begin{document}

\maketitle

\begin{abstract}
  Increasing test-time computation has emerged as a promising direction for improving language model performance, particularly in scenarios where model finetuning is impractical or impossible due to computational constraints or private model weights. However, existing test-time search methods using a reward model (RM) often degrade in quality as compute scales, due to the over-optimization of what are inherently imperfect reward proxies. We introduce \rlhfmethod, a new test-time alignment approach.  As we scale test-time compute, \rlhfmethod converges to sampling from the optimal aligned distribution for each prompt. 
  By adopting recent advances in Markov chain Monte Carlo for text generation, our method enables better-aligned outputs without modifying the underlying model or even requiring logit access. We demonstrate the effectiveness of \rlhfmethod on mathematical reasoning benchmarks ({GSM8K} and GSM-Symbolic) using a task-specific RM, showing consistent improvements over existing test-time compute methods like best-of-$n$ and majority voting. When applied with more realistic RMs trained on the \textsc{T\"{u}lu 3} preference dataset, \rlhfmethod outperforms direct preference optimization (DPO), best-of-$n$, majority voting, and weighted majority voting on a diverse range of datasets (GSM8K, MATH500, IFEval, MMLU-Redux,  and TruthfulQA).
  A practical solution to aligning language models at test time using additional computation without degradation, our approach expands the limits of the capability that can be obtained from off-the-shelf language models without further training.   
\end{abstract}

\section{Introduction}

Language models (LMs) have demonstrated remarkable capabilities through learning from human preferences  \citep{firstrlhf,fernandes2023bridging, kaufmann2023survey,10.1145/1273496.1273590,peng2019advantageweighted,rlhfandbayes,korbak2022reinforcement,go2023aligning}. However, there are a number of problems with deploying a single aligned model: alignment approaches typically average multiple human preferences, constructing a monolithic preference model and target policy.
% In reality, preferences vary dramatically across cultures, backgrounds, and belief systems and can even shift for a single individual based on context and as a function of time\gf{pluralistic alignment citation}. 
Furthermore, approaches to adapting these models through finetuning have become increasingly impractical, requiring enormous computational resources. They are entirely impossible when model weights are private, as with many state-of-the-art models \citep{openai2024gpt4technicalreport,TheC3,geminiteam2024geminifamilyhighlycapable}.% \nascomment{check bibliography so citation names render reasonably (e.g. ``Team et al'' seems wrong).  missing year on the second one}

%\alisa{I would say sampling a single generation is not about ``conventional alignment methods,'' which only pertains to training}
While conventional alignment methods optimize a single model for aggregate performance across a distribution of prompts and then produce a random generation during inference, researchers have found great improvement from scaling the amount of compute expended at test time for each prompt \citep{brown2024largelanguagemonkeysscaling,snell2024scalingllmtesttimecompute}. Here, multiple outputs are generated and then used to produce a final answer either via reward maximization (known as ``best-of-$n$,'' or BoN; \citealp{gao2022scalinglawsrewardmodel,stienon2020learning,fern2022qualityaware}), majority voting (MV; \citealp{selfconsistency}) or weighted majority voting (WMV; \citealp{li-etal-2023-making}). While promising, existing search-based methods face fundamental limitations \citep{liu2024dontthrowawayvalue,wu2024inferencescalinglawsempirical,zhang2023planninglargelanguagemodels,xie2023selfevaluationguidedbeamsearch}: as inference compute scales, these methods over-optimize learned reward models (RMs), which are inherently imperfect proxies \citep{gao2022scalinglawsrewardmodel}. %\gf{Can we add a phrase connecting to the history test-time scaling and MT? }

In this paper, we introduce \rlhfmethod, a test-time alignment method that converges to sampling from the \textbf{optimal aligned distribution} (defined in \S\ref{eq:modifiedgibbs} as the true target for existing RLHF methods) for each prompt, individually, as the test-time compute budget is increased. This enables more aligned responses without requiring access to the logits or training any underlying LM (only sampling from it). Specifically, we adapt \method \citep{quest}. \method was designed to repurpose a language model, creating a Markov chain converging toward a (Gibbs) distribution defined by machine translation quality estimation metrics. In this work, we show that we can use \method to align LMs at test time according to an RM learned from preference data (in place of the quality estimation metric) and to provide a single good final prediction.

\begin{figure}[t]
    \centering
    \includegraphics[width=0.65\textwidth]{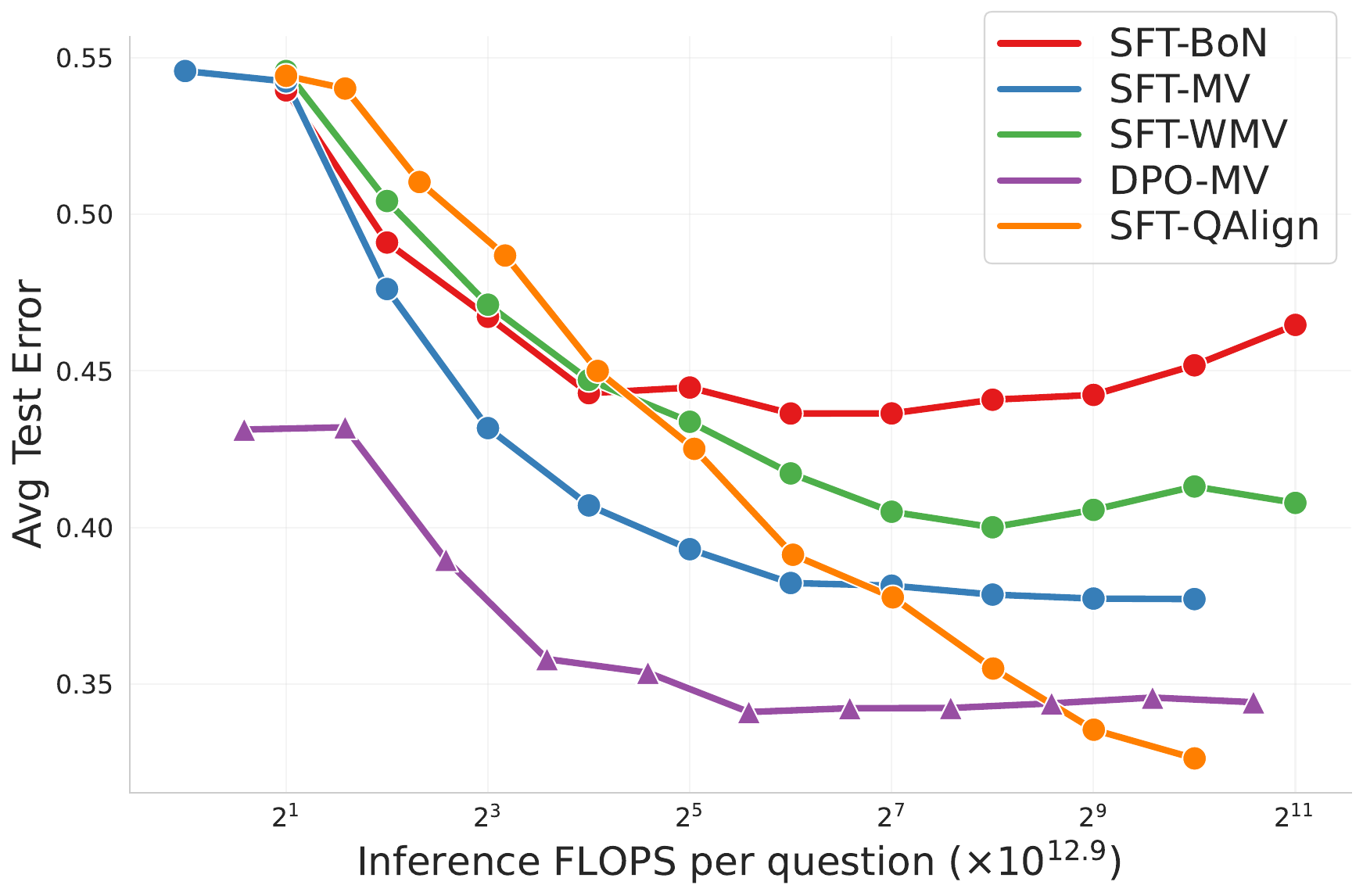}
    \caption{Average error rate across multiple evaluation datasets (GSM8K, MATH500, MMLU-Redux, TruthfulQA, and IFEval) as a function of inference-time floating point operations (FLOPS) in log scale. 
    We compare \textbf{\textcolor{SFT-QUEST}{\raisebox{-0.1em}{\scalebox{1.7}{$\bullet$}}\rlhfmethod with \textsc{T\"ulu3-8B-SFT}}}  against four baselines: \textbf{\textcolor{DPO-MV}{\raisebox{0.0em}{\scalebox{1.2}{$\blacktriangle$}} majority vote (MV) \textsc{T\"ulu3-8B-DPO}}}, and applied to \textsc{T\"ulu3-8B-SFT} the methods \textbf{\textcolor{SFT-BoN}{\raisebox{-0.1em}{\scalebox{1.7}{$\bullet$}} best-of-$n$ (BoN)}},  \textbf{\textcolor{SFT-MV}{\raisebox{-0.1em}{\scalebox{1.7}{$\bullet$}} MV}}, and \textbf{\textcolor{SFT-WMV}{\raisebox{-0.1em}{\scalebox{1.7}{$\bullet$}} weighted MV (WMV)} }. All experiments use temperature 1.0 with reasoning included in model outputs. The \textsc{T\"ulu3-8B-DPO} model results from preference finetuning \textsc{T\"ulu3-8B-SFT} (approximately $1.75 \times 10^{19}$ FLOPs). The costs of this process are not accounted for in this plot. 
    }\label{fig:questoutperform}
\end{figure}
In our experiments (\S\ref{sec:experiment}), we compare \rlhfmethod with existing test-time compute methods on mathematical reasoning benchmarks (\textbf{GSM8K}, \citealp{gsm8k}, and \textbf{GSM-Symbolic}, \citealp{gsmsymb}), by training a task-specific RM (\S\ref{sec:task-specific}), and using it to increase the capability of \textsc{Llama3.1-8B-Instruct} \citep{llama3}. We obtain a consistent reduction in error as we scale inference time computation across scales, as shown in Figure~\ref{fig:tasktuning_accscores}. Unlike BoN, \rlhfmethod's performance does not degrade as we increase the compute per question.  
%\gf{delete if not in the plot, outperforming \textsc{Llama3-405B-Instruct} \citep{llama3}}
Furthermore, when applied with more realistic RMs (\S~\ref{sec:general-exp}), trained on the \textsc{T\"{u}lu3} \citep{tulu3} preference dataset, where over-optimization starts to appear at a lower inference budget, and applied to \textsc{T\"{u}lu3-8B-SFT}, our experiments show that \rlhfmethod can consistently outperform direct preference optimization (DPO; \citealp{dpo}), which updates the model weights, as well as BoN, MV, and WMV on a diverse range of datasets in mathematical reasoning (\textbf{GSM8K} and \textbf{MATH500}; \citealp{hendrycksmath2021,lightman2023letsverifystepstep}), knowledge recall (\textbf{MMLU-Redux}; \citealp{gema2025mmlu}), \textbf{TruthfulQA} \citep{truthfulqa} and instruction following (\textbf{IFEval}; \citealp{ifeval}), as shown in Figure~\ref{fig:questoutperform}.\footnote{Website with code and additional resources will be provided upon acceptance.}%\url{w

Our contributions are:
\begin{itemize}
    \item We propose \rlhfmethod, a method for making local approximations of the optimal aligned distribution at test time.
    \item We show a {$28.4\%$} increase in accuracy on GSM8K and {$78.34\%$} on GSM-Symbolic relative to a single generation, using a task-specific RM, beating BoN, MV, and WMV.
    \item We show that \rlhfmethod outperforms DPO when applied to \textsc{T\"{u}lu3} family of models, in general RMs trained on the \textsc{T\"{u}lu3} preference dataset, even when compared with the same compute budget at test-time via MV, achieving a $47.85\%$ increase in average accuracy relative to a single generation across a suite of benchmarks.
\end{itemize}

%\newpage

\section{Background: Language Model Alignment}
\label{sec:background}

%Generating text from autoregressive LMs involves conditioning on a prompt $x$ encoding the particular question we want to solve. 
%Let $\mathcal{Y}$ denote the set of possible LM responses (outputs).
%The  distribution over responses $y \in \mathcal{Y}$, given the prompt $x$, can be factorized as the product of conditional probabilities over individual tokens $\langle y_1, y_2, \ldots, y_N\rangle$, where each $y_i \in \mathcal{V}$, a fixed vocabulary. The probability of a particular response $y$ for a given input $x$ can be written as
%\begin{equation}
%    p_{\text{LM}}(y \mid x) = \textstyle \prod_{i=1}^{N} p_{\text{LM}}(y_i \mid y_{<i}, x).
%\end{equation} Here, $y_{<i} = \langle y_1,y_2, \ldots, y_{i-1}\rangle$.%  \nascomment{should the conditional above be $p_{\text{LM}}(y_i \mid x, y_{<i})$?  or even $p_{\text{LM}}(y_i \mid x \oplus y_{<i})$ to show concatenation?}\gf{feel its the same and this looks more pleasant}

Finetuning a pretrained LM $p_\text{LM}(y \mid x)$  to align with preferences encoded by a reward function $r(y,x)$ can be cast as Bayesian inference \citep{korbak2022reinforcement}. We seek $\pi(y\mid x)$, which is initially equal to a prior $p_\text{LM}(y \mid x)$, and updated to conform to evidence provided by the human preferences model. Let $\gamma=1$ denote the event that response $y$ is maximally preferred for prompt $x$ (i.e., $y \succeq y'$ for all $y' \in \mathcal{Y}$). Assuming the reward function is bounded, we can model the likelihood of this event as $p(\gamma=1 \mid y,x) = \exp\left(\left({r(y,x) - \max_{y'} r(y',x)}\right)/{\beta}\right)$. Using Bayes' rule, the posterior takes the following form:
\begin{equation} \label{eq:modifiedgibbs}
\pi(y \mid x) \triangleq p(y \mid \gamma=1,x) = \textstyle \frac{1}{Z_\beta(x)}p_\text{LM}(y \mid x)\exp \left(\frac{r(y, x)}{\beta} \right).
\end{equation}
It is both intractable to compute the partition function $Z_\beta(x) = \sum_{y\in\mathcal{Y}} p_\text{LM}(y\mid x)\exp(r(y, x)/\beta)$, and to sample from the distribution in Eq.~\ref{eq:modifiedgibbs} exactly.

To approximate $\pi(y\mid x)$, many methods have been proposed \citep{firstrlhf,fernandes2023bridging, kaufmann2023survey,10.1145/1273496.1273590,peng2019advantageweighted,rlhfandbayes,korbak2022reinforcement,go2023aligning}. Given a dataset of prompts $\mathcal{D}$, one common approach is to estimate a parameterized variational approximation $q_\theta(y\mid x)$(see Appendix~\ref{appendix:rlhfvi} for details):
\begin{equation}
\max_\theta \mathbb{E}_{x\sim\mathcal{D}} \big[ \mathbb{E}_{y \sim q_\theta(y\mid x)}\left[ r_\phi(y,x)/\beta \right] - D_{\text{KL}} \left( q_\theta(y\mid x)\ \|\ p_{\text{LM}}(y\mid x) \right) \big].\label{eq:rlhfobj}
\end{equation}
%\nascomment{$\mathcal{D}$ is never explained.  also not clear what $q$  has to do with $\pi$, which is what we said we really want.  Is the idea that $q$ is the approximation of $\pi?$  the other thing that's a bit unclear is how updating $\theta$ relates back to updating the parameters of the LM}\gf{ What do you mean? $\theta$ are the parameters of the LLM, $q$ starts as the LLM $p_\text{LM}$ and then is optimized via policy gradient methds to maximize the lower bound.} \gf{I maid it clear that q is an LM that is initialized with the base model.}
Methods like PPO \citep{schulman2017proximalpolicyoptimizationalgorithms} and others \citep{shao2024deepseekmathpushinglimitsmathematical,dpo,rloo} optimize this objective with low-variance gradient estimates, resulting in stable optimization. Importantly, $q_\theta(y\mid x)$ is a newly trained LM  initialized with $p_\text{LM}(y\mid x)$.% The large-scale efforts to align language models generally alternate between learning $q_\theta(y\mid x)$ and $r(y,x)$, with human preferences on samples generated from $q_\theta(y\mid x)$ \citep{llama3,llama2,geminiteam2024geminifamilyhighlycapable}. This avoids allowing the approximate posterior to exploit regions where the reward is out of distribution and also to refine the resolution of the reward towards promising regions. 

\section{Test-Time Alignment via MCMC}
\label{sec:mcmc}
Casting language model alignment as posterior inference decouples our target goal from the procedure used to achieve it. While current approaches rely on variational inference to learn a single model $q_\theta(y \mid x)$ (Eq.~\ref{eq:rlhfobj}), this strategy faces four limitations. First, it requires expensive model finetuning, which is costly with large models.  Second, many models' weights are not openly shared, including those of  state-of-the-art models like GPT-4 \citep{openai2024gpt4technicalreport} and Gemini \citep{geminiteam2024geminifamilyhighlycapable}. Third, amortizing the approximation across all prompts necessarily sacrifices the quality of approximation for any individual prompt $x$ to achieve good average performance. Fourth, the approach assumes a monolithic notion of human preferences encoded in $r(y,x)$, offering no flexibility to adapt to varying user preferences or contexts at inference time.

These limitations motivate a shift toward local posterior approximations, where we achieve a better approximation as we increase the compute budget at test time on a single prompt $x$. As we will describe in (\S\ref{sec:mcmc}), our key insight is that rather than estimating a single parametric approximation, we can use recent advancements in Markov chain Monte Carlo (MCMC) sampling with LMs \citep{quest} and obtain a sequence of samples $\mathcal{S} = \langle y^0, y^1, \dots, y^T \rangle$ from $\pi(y \mid x)$ and use them for \textbf{optimal decision-making}.

A principled way to select a final response from the generated candidate set $\mathcal{S}$ from $\pi(y \mid x)$ is by selecting the most common element (mode) i.e., via majority voting (MV; \citealp{selfconsistency}). However, this only works for tasks with well-defined answers, such as mathematical reasoning or multiple-choice questions. For open-ended generation tasks, we apply a generalization of this approach through the minimum Bayes risk (MBR) principle \citep{10.3115/1118693.1118712,eikema-aziz-2020-map,farinhas2023empiricalstudytranslationhypothesis,bertsch2023itsmbrwaydown}. The MBR framework selects the output $\hat{y}$ that maximizes the expected task-specific utility $u(y,y')$:
\begin{equation}
\hat{y} = \underset{y \in \mathcal{S}}{\arg \max} \;\mathbb{E}_{y' \sim \pi (y' \mid x)} \big[ u(y,y') \big] \approx \underset{y \in \mathcal{S}}{\arg \max} \frac{1}{T+1} \textstyle \sum^{T}_{t=0} u(y,y^{t}).
\label{eq:mbr}
\end{equation}
When the utility metric $u(y, y')$ is defined as an exact match over the final answer, MBR amounts to a MV strategy over $\mathcal{S}$. We employ ROUGE \citep{lin-2004-rouge} as our utility metric for open-ended generation tasks, following \cite{bertsch2023itsmbrwaydown}. While this results in $O(T^2)$ similarity computations, ROUGE is computationally lightweight. %, \gf{remove the second part of the sentence} and efficient low-rank approximations \citep{trabelsi2024efficientminimumbayesrisk} can be used when dealing with heavier neural-based metrics.

%\gf{
In the following sections, we describe two practical approaches to obtain a final answer based on MBR to samples from $\pi(y \mid x)$. First, \rlhfmethod (\S\ref{sec:mcmc}), our novel MCMC-based approach to sample from the aligned distribution $\pi(y \mid x)$ at test time, allowing us to apply MBR directly to these samples. Second, in \S\ref{sec:is}, we present importance sampling, a classical alternative that circumvents direct sampling from the aligned distribution by reweighting samples from the base model $p_{\text{LM}}(y \mid x)$ to approximate MBR as if they were drawn from $\pi(y \mid x)$. Finally, in \S\ref{sec:bon}, we describe best-of-$n$ sampling (BoN), a simple and widely used baseline approach for test-time alignment that does not explicitly optimize an MBR objective but instead selects a single high-reward sample from the base model.
%}

%In \S\ref{sec:mcmc}, we introduce \rlhfmethod, our novel MCMC-based approach to sample from the aligned distribution $\pi(y \mid x)$ at test time. \nascomment{something like:  ``We then turn to a view of existing test-time alignment strategies through the MBR lens.''} In \S\ref{sec:is}, we present importance sampling, a classical alternative that circumvents direct sampling from the aligned distribution by estimating the expectation in Eq.~\ref{eq:mbr} by reweighting samples from the base model $p_\text{LM}(y \mid x)$. Finally, in \S\ref{sec:bon}, we describe best-of-$n$ sampling (BoN), a simple and widely used baseline approach for test-time alignment that does not explicitly optimize an MBR objective but instead selects a single high-reward sample from the base model.

%that we can use to make  final decisions \gf{(\S~\ref{sec:optimaldecision})}.
%In section, 
% In essence, instead of approximating the target density in Eq.~\ref{eq:modifiedgibbs}, we will simulate generating from it for each prompt $x$.

\subsection{MCMC for Text Generation } 
\label{sec:mcmc}
% In section \ref{sec:mcmc}, we introduce our approach that applies MCMC to better approximate this expectation. In section \ref{sec:importancesampling} we introduce a broad family of approaches to try to approximate this expectation and show how different sampling methods in the literature are variants of it.

%\subsection{MCMC for Text Generation } 
%\label{sec:mcmc}

% Like importance sampling (\S)\ref{sec:importancesampling}),
% However, unlike importance sampling, we will not use the distribution $p_\text{LM}(y \mid x)$ to obtain a set  of independent samples. 
% MCMC will sample from a proposal distribution. 
% \alisa{I suggest starting with what you actually do, and then making these comparisons later if relevant. right now this first paragraph is delaying getting to the important part.}\gf{great point noted}

With the goal of generating a sequence of samples from $\pi_\beta(y \mid x)$, we will construct a Markov chain $(y^0, y^1, \ldots, y^T)$ that has $\pi_\beta(y \mid x)$ as its equilibrium distribution. The chain starts from a hypothesis $y^0 \sim p_\text{LM}( y \mid x)$. On the $t$th iteration, it draws a new hypothesis $y$ from a \textbf{proposal distribution}  $q(y \mid y^{t}, x)$ , and this hypothesis is accepted with an acceptance probability $\alpha_\beta(y, y^t) \in [0,1]$. The \textbf{proposal} distribution $q(y \mid y^{t}, x)$ we use is the one proposed by \method \citep{quest}. It is based on a LM that samples suffixes $p_\text{LM}(y_{i:N} \mid y_{<i}, x)$ starting at uniformly sampled index $i$. This proposal can be written as
\begin{equation}
q(y\mid y^{t},x,i) = p_{\text{LM}}( y_{i:N} \mid  y^{t}_{<i},x) \times \boldsymbol{1}\{y_{1:i} = y_{1:i}^t\},
%\textstyle \prod_{j < i} \delta(y_j ,y^{t}_j), 
\label{eq:proposal}
\end{equation}
%where $\delta(y_j ,y^{t}_j)$ is the Kronecker delta function, which assigns zero  to prefix tokens which are different from $y^t_{<i}$ and  one to tokens matching the prefix. 
Following the Metropolis-Hastings algorithm (MH; \citealp{hastings}), the acceptance probability is 
\begin{equation}\label{eq:acceptance}
\alpha_\beta(y, y^t) = \min \left\{1, \,\, \left. \pi_\beta \left(y\mid x \right)q \left(y^{t} \mid y,x \right) \ \right/\ 
{\pi}_\beta \left(y^{t} \mid x \right)q\left(y \mid  y^{t},x\right)\right\}.
\end{equation}
If the candidate $y$ is accepted, the next state in the chain becomes $y^{t+1} = y$; if rejected, the chain stays at $y^{t+1} = y^t$. The process repeats for some number of steps $T$.  In the end, it returns the set of accepted samples. Note that, while computing the likelihood $\pi_\beta(y\mid x)$ of a particular hypothesis $y$ under the target distribution is intractable (due to the  partition function $Z_\beta(x)$), evaluating the acceptance criterion $\alpha_\beta(y, y^t)$ is easy, because it  depends only on the likelihood ratio, in which the normalization constants cancel out: 
\begin{equation}\label{eq:targetration}
\frac{{\pi}_\beta(y\mid x)}
{{\pi}_\beta(y^{t} \mid x)} = \exp \left( \textstyle \frac{r\left(y,x\right) - r\left(y^t,x\right)}{\beta} \right) \displaystyle \frac{p_\text{LM}( y \mid x)}{p_\text{LM}( y^{t} \mid x)}.
\end{equation}
MH converges to the unique stationary distribution $\pi_\beta(y \mid x)$, regardless of the initial distribution, because the transition distribution of the Markov chain,  $p(y^t \mid y^{t-1},x)$ which results from generating a candidate from $q(y^t \mid y^{t-1}, x)$ followed by an accept/reject step, satisfies the \textit{Markov chain ergodic theorem} \citep{Neal2011ProbabilisticIU}(see Appendix~\ref{sec:convergenceproof} for a detailed proof). 

Note that, under samples from the proposal from Eq.~\ref{eq:proposal}, the likelihood ratio from Eq.~\ref{eq:targetration}  is proportional to the inverse of the probability of returning:
\begin{equation}
\frac{q(y^t\mid y,x,i)}{q(y \mid y^t,x,i)} = \frac{p_\text{LM}( y^{t}_{i:N} \mid  y^{t}_{<i},x) }{p_\text{LM}( y^{}_{i:N} \mid y^{t}_{<i},x)},
\end{equation}
this allows simplifying the criterion as:\begin{equation}\label{eq:rlhfcretirion}
\alpha_\beta(y, y^t) = \min \left\{ 1, \exp \left(\textstyle\frac{ r\left(x,y\right) - r\left(x,y^t \right)}{\beta} \right)  |y^t|/ |y| \right\}.
\end{equation}
The length ratio ${|y^t|}/{|y|}$ comes from the ratio between the uniform index distributions.  
Note that this means that we can sample from the aligned distribution in Eq.~\ref{eq:modifiedgibbs} without any access to logits or parameter weights.

In summary, we propose a simple and effective procedure for sampling from $\pi(y \mid x)$. We characterize the \rlhfmethod sampling process as repeating the following for $T$ steps:
\begin{enumerate}
    \item Given an instance $y^t$ with length $|y^t|$, sample an index $i$ uniformly.
    \item Generate a completion $y_{i:N}$ from $p_\text{LM}(y_{i:N} \mid y^{t}_{<i}, x)$.
    \item Compute the probability of acceptance on the reward difference in Eq.~\ref{eq:rlhfcretirion}. Sample a random boolean based on this probability. If we reject, $y^{t+1} = y^t$; if we accept,  $y^{t+1} = y'$.
\end{enumerate}

The \method proposal is simple and relies solely on the model's ability to generate text left-to-right, with the consequence that successive samples are only conditionally dependent up to the index $i$, which limits the speed at which we can explore (i.e., the effective sample size).

While more complex proposals could be built based on LMs prompted to self-refine \citep{solvingchallengingmathword,selfrefine, react}, recent work  demonstrates that even strong LMs often fail to improve reasoning due to models' inability to gauge the correctness of their outputs and localize errors \citep{cantselfcorrect}. Because of this, a growing body of research tries to teach LMs how to self-correct by creating data-augmentation recipes \citep{welleckselfcorrect,glore} or new datasets \citep{wang2023shepherdcriticlanguagemodel, chen2024improvingcodegenerationtraining, lee2024platypusquickcheappowerful, saunders2022selfcritiquingmodelsassistinghuman, schick2022peercollaborativelanguagemodel}. However, when the proposal is different from the base model, we lose the simplicity of the acceptance criterion in Eq.~\ref{eq:rlhfcretirion} and need to keep track of two sets of logits to calculate the acceptance criterion in Eq.~\ref{eq:acceptance}.

As our approach only requires access to the ability to generate continuations given a prefix (not access to base model parameters or logits), it can also be used with closed LMs through an API.  However, we limit our experiments to open-weight models. We provide in Appendix~\ref{sec:computation} an analysis of the computational cost associated with running \rlhfmethod. 

\subsection{Importance Sampling}\label{sec:is}

%\gf{salient the novelty of this connection?}
A natural competing approach to \rlhfmethod for approximating the intractable expectation in Eq.~\ref{eq:mbr} is to use importance sampling (IS; \citealp{is}). Rather than attempting to directly sample from the target distribution $\pi(y\mid x)$, IS generates samples from the base LM $p_\text{LM}(y\mid x)$ and reweights them to match the target distribution $\pi(y\mid x)$:
\begin{align*}
\mathbb{E}_{y \sim \pi(y\mid x) } [ u(y,y') ] &= \mathbb{E}_{ y \sim p_{\text{LM}}(y\mid x)} \left[\frac{ \pi(y\mid x) }{p_{\text{LM}}(y\mid x)} u(y,y')   \right]  
= \mathbb{E}_{ y \sim p_{\text{LM}}(y\mid x)} \left[\frac{ \exp( \frac{1}{\beta}r(y,x)) }{ Z_\beta(x) } u(y,y')\right].
\end{align*}
The partition function $Z_\beta(x)$ can be reframed as an expectation over samples from the base model: $Z_\beta(x) = \sum_{y\in\mathcal{Y}} p_\text{LM}(y\mid x)\exp(r(y, x)/\beta) = \mathbb{E}_{y \sim p_\text{LM}(y\mid x) } [\exp(r(y, x)/\beta) ]$. This motivates a self-normalizing importance sampling approach, where we approximate both the numerator and denominator using samples $\{y^{(i)}\}_{i=1}^K \sim p_\text{LM}(y\mid x)$. The self-normalized estimator is consistent, but introduces bias due to the correlation between the numerator and denominator \citep{owen2013monte}. In the end, making this approximation boils down to the following steps:
\begin{enumerate}
\item Sample $y^{(0)} , \ldots , y^{(T)} \sim p_\text{LM}(y\mid x)$. 
\item Evaluate the generations with the reward model $r(y,x)$ resulting in $r^{(0)} , \ldots , r^{(T)}$. 
\item Obtain the importance weights $w^{(i)} = \exp( r^{(i)} / \beta) \ \left/\  \textstyle\sum_{j=0}^{T}  \exp( r^{(j)} /\beta)\right.$.
\item For each hypothesis $y'$,   compute $\mathbb{E}_{y \sim \pi(y \mid x) } [ u(y,y') ]  \approx \sum_{i=0}^{T} w^{(i)}  u(y,y^{(i)})$.    
\end{enumerate}

In the LM literature, this procedure, for the specific case of tasks with well-defined final answers (i.e., the utility metric $u(y, y')$ is defined as an exact match), the MBR output is the same as \textbf{weighted majority voting}  (WMV; \citealp{li-etal-2023-making}). Similar to \rlhfmethod, WMV is guaranteed to converge to the optimal decision rule as computational resources increase. However, because it relies solely on independent samples drawn from the base LM, the generation process cannot be directly steered toward more promising regions of the output space. This limitation becomes particularly problematic when the base LM infrequently produces high-reward responses. In such cases, the approximation of the target expectation, and corresponding final decision, will be compromised.
%
%\nascomment{we never explain MV through this lens}\gf{we do know whem we say MV is exact match on u.}

\subsection{Best-of-$n$ Sampling}
\label{sec:bon}
%\nascomment{again, need more scaffolding here to situate this as a competing method.  is BON motivated by any particular MBR type goal?  whether yes or no, say so}

Best-of-$n$ (BoN) sampling is a simple approach for aligning language model predictions at test time. BoN has emerged as a strong baseline in the literature, requiring no additional training, while still achieving compelling performance. Unlike \rlhfmethod\ and IS, which explicitly optimize for a MBR objective, BoN uses a heuristic selection process.

Given a prompt $x$, we generate $n$ candidate responses $y_1,y_2, \ldots,y_n$ independently from the base LM $p_\text{LM}(y \mid x)$. Each response is evaluated using the RM $r(x,y)$, and the candidate with the highest reward is chosen as the final output:
\begin{equation}
y^\ast(n) = \textstyle \arg\max_{y_i} r(x,y_i).
\end{equation}
This procedure implicitly biases the output distribution toward higher-reward responses. It is easy to see that in the limit of $n$, we are sampling from the maximum reward distribution; however, BoN is remarkably performant even when  $n$ is small \citep{fern2022qualityaware,gao2022scalinglawsrewardmodel,llama3}.

With some strong assumptions, we can show that the probability density of the BoN distribution can be approximated by our target distribution $\pi(y \mid x)$ from Eq.~\ref{eq:modifiedgibbs} in the special case when $\beta$ takes the form:
%the following form:
%\begin{equation}p(y^\ast(n) \mid x) \propto \exp\!\left(\frac{2(n-1)\, r(x,y^\ast(n))}{\sigma_r(x)\sqrt{2\pi}}\right)\, p_{\text{LM}}(y^\ast(n)\mid x),
%\begin{equation}p(y^*(n) \mid x) \propto \exp\left( r(x,y^*(n)) \:/\: \beta^*(n) \right) \, p_{\text{LM}}(y^*(n) \mid x),
%\end{equation} where $\beta$ takes the form
 \begin{equation}
    \beta(n) = \frac{\sigma_d(x)}{\sqrt{2\log n_d} - \frac{(\log\log n_d + \log(4\pi))}{2\sqrt{2\log n_d}}}.
\end{equation} where $\sigma_d(x)$ is the standard deviation of the dominant component of the distribution of reward, under the assumption of Gaussian mixture, which we empirically observe (see Appendix~\ref{eq:empiricaldist}), when evaluated on samples from our base LM conditioned on $x$, and $n_d=w_d(x) n$, the dominant component's sample size, where $w_d(x)$ is the weight of the dominant component. A full derivation of this approximation is provided in Appendix \ref{appendix:bonproof}. % Note that this density equals our target distribution $\pi(y \mid x)$ from Eq.~\ref{eq:modifiedgibbs} if we set $\beta =\beta^\ast(n)$.

This rough equivalence, not noted in past literature to our knowledge, provides a new insight into a key distinction between BoN and the other methods we have discussed, including \rlhfmethod. In \rlhfmethod, we explicitly define a target distribution $\pi(y \mid x)$, and increasing the test-time budget $n$ refines our approximation of this fixed objective. In contrast, BoN does not converge to a predefined distribution; instead, increasing $n$ progressively shifts the selection process toward responses with higher rewards and deviating further from $p_{\text{LM}}(y \mid x)$. Insofar as the reward model is well-aligned with human preferences, this should make BoN highly effective. However, in realistic settings where the reward is imperfect, increasing $n$ can lead to over-optimization, i.e.,  selecting responses that score well under the reward model but degrade in actual quality. As a result, BoN can suffer from diminishing returns or even performance degradation as test-time compute increases, which we observe in our experiments (\S\ref{sec:task-specific}--\ref{sec:general-exp}) and previous work \citep{gao2022scalinglawsrewardmodel}. % \gf{ speculation->we could make an estimator called BON resampler that from N independent samples we resample a few times the top K - tailored to a value $\beta$. and than do MBR on that. I would assume this would converge to the WMV estimator. I have experiments showing this is the case but seems a bit out of scope. }\gf{This is only partially true, because while it can match the mean, it will have a lower variance than the underlying $\pi(y\mid x)$}

\section{Experiments} \label{sec:experiment}

Our experiments are centered around two main questions.  First, \textbf{task-specific tuning:} given a powerful task-specific RM trained on preference data derived from ground-truth evaluations of base model outputs, can \rlhfmethod outperform BoN and WMV (\S\ref{sec:task-specific})? %\gf{ add robustness to distribution shifts ?}
Second,  \textbf{general alignment:} when applied as an alternative to DPO, where the preference dataset encompasses multiple tasks aimed at improving the instruction-following capabilities of chat models, can \rlhfmethod outperform state-of-the-art alignment methods (\S\ref{sec:general-exp})?

\subsection{Task-Specific Tuning}
\label{sec:task-specific}

We evaluate \rlhfmethod for task-specific tuning using \textsc{Llama-3.1-8B-Instruct} \citep{llama3} as the base model and a custom \citet{bt} RM trained on our preference dataset for mathematical reasoning (training details in Appendix~\ref{appendix:rmtraining}). This RM was initialized with \textsc{Llama-3.1-8B-Instruct} and trained on 64 model-generated response pairs per prompt.  We selected pairs  based on ground truth answers.\footnote{Data and RM will be provided upon publication.}

\paragraph{Baselines}  We compare against BoN, and WMV sampling applied to \textsc{Llama-3.1-8B-Instruct} using our trained RM and MV from samples from \textsc{Llama-3.1-8B-Instruct}. For each of the methods we generate $1024$ solutions per problem.

\paragraph{Datasets}  We evaluate on \textbf{GSM8K} \citep{gsm8k}, a dataset of grade school math problems, and \textbf{GSM-Symbolic} \citep{gsmsymb}, a new benchmark designed to assess out-of-distribution generalization in mathematical reasoning. As demonstrated by \citet{gsmsymb}, GSM-Symbolic exposes substantial performance degradation across state-of-the-art LLMs exhibiting drops of up to $65\%$ compared to their GSM8K scores when faced with simple variations like changed numerical values and extra redundant clauses.
% drops of up to 65% across all state-of-the-art models

%a verified subset of the MATH  test dataset that contains advanced mathematical problems across algebra, geometry, and calculus \citep{hendrycksmath2021}.
%We evaluate on \textbf{GSM} \citep{gsm8k}, a dataset of grade school math problems, and \gf{GSM-symb}\textbf{MATH500} \citep{lightman2023letsverifystepstep},  a verified subset of the MATH  test dataset that contains advanced mathematical problems across algebra, geometry, and calculus \citep{hendrycksmath2021}.
\paragraph{Sampling configurations} We sampled both datasets with temperature $1.0$ and $\beta=1$. Since the acceptance rate is a function of $\sigma_r(x)/\beta$ (where $\sigma_r(x)$ is the distribution of rewards under the base model), we selected this value to achieve approximately $50\%$ acceptance rate based on tuning with 128 samples from the GSM8K training data. This relatively high acceptance rate ensures the chain mixes well and avoids getting stuck in a single mode.

\paragraph{Results}

\begin{figure}[h]
    \centering
    \begin{subfigure}[b]{0.49\textwidth}
        \centering
        \includegraphics[width=\textwidth]{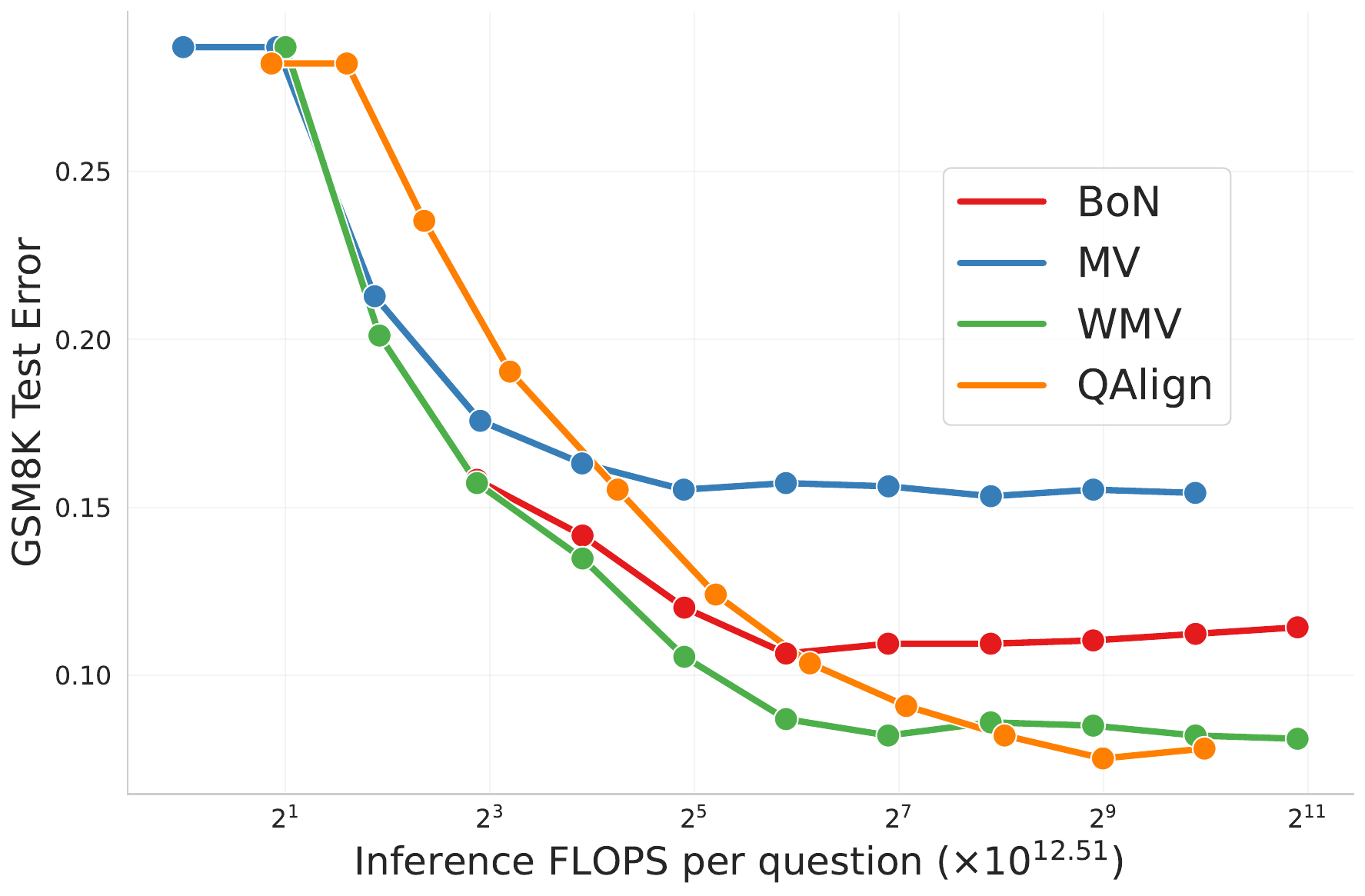}

    \end{subfigure}
    \hfill
    \begin{subfigure}[b]{0.49\textwidth}
        \centering
        \includegraphics[width=\textwidth]{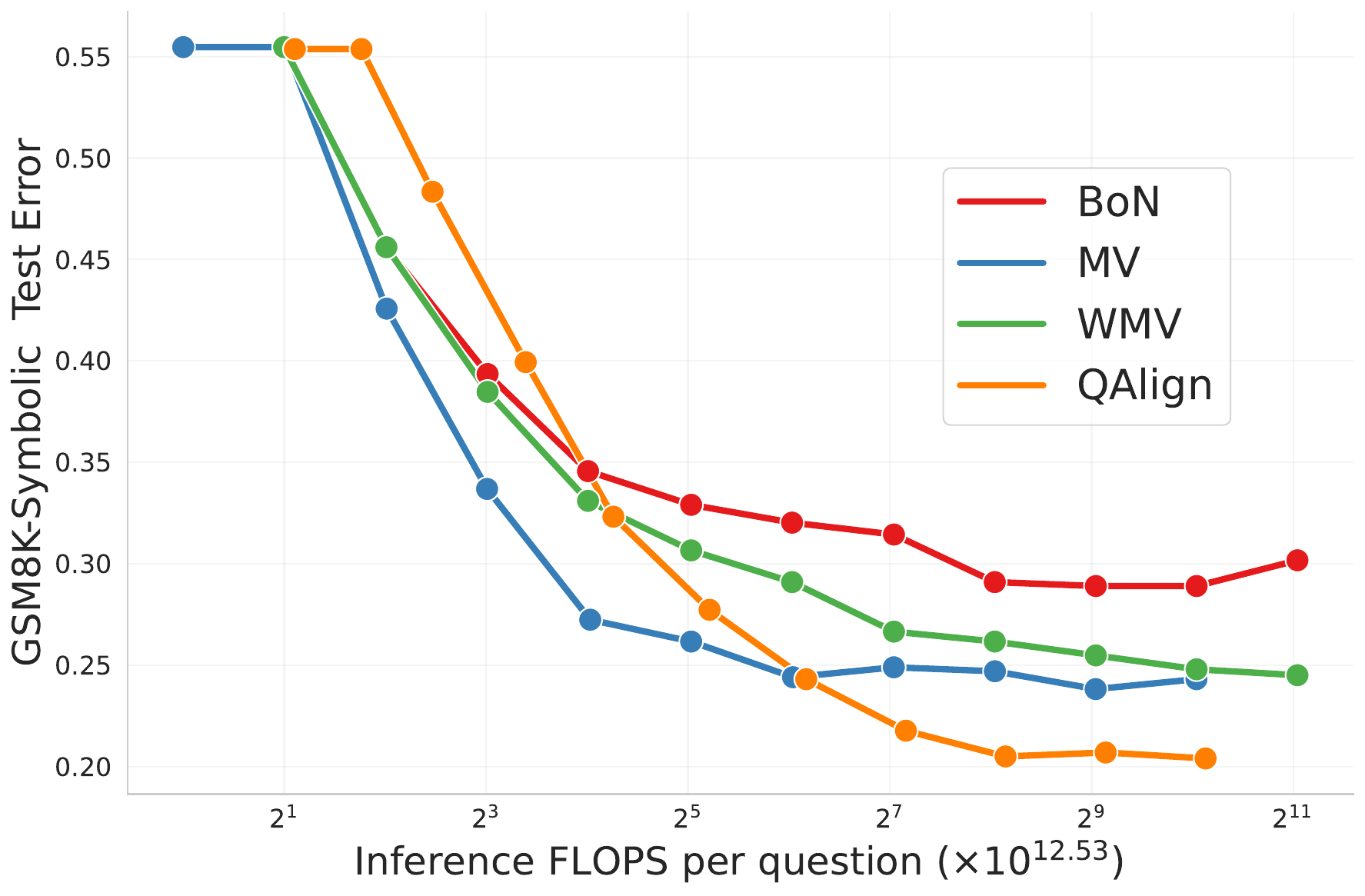}
    \end{subfigure}
    \caption{Average accuracy vs.~floating point operations (FLOPS) in log scale. 
    We compare \textbf{\textcolor{SFT-QUEST}{\raisebox{-0.1em}{\scalebox{1.7}{$\bullet$}} \rlhfmethod with \textsc{Llama-3.1-8B-Instruct}}} against three baselines also applied to \textsc{Llama-3.1-8B-Instruct}: \textbf{\textcolor{SFT-BoN}{\raisebox{-0.1em}{\scalebox{1.7}{$\bullet$}} best-of-$n$ (BoN)}}, \textbf{\textcolor{SFT-MV}{\raisebox{-0.1em}{\scalebox{1.7}{$\bullet$}} majority vote (MV)}}, and \textbf{\textcolor{SFT-WMV}{\raisebox{-0.1em}{\scalebox{1.7}{$\bullet$}} weighted MV (WMV)}}. 
    \textit{Left}: Error rate (lower is better) on GSM8K test dataset. \textit{Right}: Error rate on GSM-Symbolic test dataset. All experiments use temperature 1.0 with reasoning included in model outputs.}
    %\caption{Average accuracy vs.~floating point operations (FLOPS). We evaluate the error rate (lower is better) of an increasing number of sampled solutions for majority voting (MV), best-of-$n$, and \rlhfmethod. \textit{Left}: GSM test dataset. \textit{Right}:  MATH test dataset. The FLOPS axis is in log-scale. \nascomment{why no MV on right?} \gf{serious issue with MATH rm. working on it. }}
    \label{fig:tasktuning_accscores}
\end{figure}

The results of the task-specific tuning experiments are summarized in Figure \ref{fig:tasktuning_accscores}. \rlhfmethod shows progressively lower error rates across both problems as we spend more computational resources. In contrast, on GSM8K, MV shows initial improvement but quickly saturates, while BoN displays improvement with additional compute, temporarily outperforming \rlhfmethod, but eventually reaches an inflection point, after which error rates begin to increase. This behavior aligns with the observations from \cite{gao2022scalinglawsrewardmodel} and theoretical results from \S\ref{sec:bon}. 
%As BoN increases the number of samples, it over-optimizes the learned RM, which is an imperfect proxy for the target metric and can thus lead BoN astray, causing it to exploit the RM's limitations rather than improving on the true objective. 
WMV performs well on GSM8K but becomes relatively worse after a budget of approximately $2^{8}\times 10^{12.51}$ FLOPs.

While all methods show some performance drop on GSM-Symbolic compared to GSM8K, the magnitude varies significantly. BoN and WMV, which rely on the RM, struggle more with the distribution shift, likely because the RM is fit to particular GSM8K-specific patterns. Relative to these two, MV shows greater robustness to these changes. This is primarily because under the distribution shift the RM becomes less reliable, and since MV does not use it, MV's performance degrades less. However, even under these circumstances, \rlhfmethod is able to leverage the RM and outperform MV on this dataset with enough compute. This suggests that \rlhfmethod is robust to imperfections in the RM and able to extract a useful signal even when the RM's reliability is compromised.

% Using \rlhfmethod, we demonstrate that with \rlhfmethod and \textsc{Llama-3.1-8B-Instruct} we can achieve or exceed the performance of \textsc{Llama-3.1-405-Instruct} across both datasets (shown in Figure~\ref{fig:tasktuning_accscores}). \nascomment{left plot shows 70B not 405B.}\gf{I have to reproduce the results of this of larger models.}\gf{add wmv comments}\gf{should I report their values for the paper where they do 8shot and stuff ?}

\subsection{General Alignment}
\label{sec:general-exp}

We evaluate \rlhfmethod for more general alignment using \textsc{T\"ulu3-8B-SFT} as the base model and \textsc{T\"ulu3-8B-RM} as the RM. The \textsc{T\"ulu3} model family \citep{tulu3} was selected for being fully open source, providing instruction-tuned, and aligned versions, along with their corresponding training code and datasets. 
This enables direct comparison of \rlhfmethod on top of the instruction-tuned model to the aligned model, sharing the initial starting point. Additionally, they provide an RM trained on the data used to train the DPO model, which we use in our experiments.

\paragraph{Baselines} We compare against 
(1) MV applied to \textsc{T\"ulu3-8B-DPO}, (2) BoN sampling applied to \textsc{T\"ulu3-8B-SFT} using \textsc{T\"ulu3-8B-RM}, (3) WMV applied to \textsc{T\"ulu3-8B-SFT} using \textsc{T\"ulu3-8B-RM}, and (4) MV applied to \textsc{T\"ulu3-8B-SFT}. Note that \textsc{T\"ulu3-8B-DPO} model is the result of doing preference finetuning on the \textsc{T\"ulu3-8B-SFT} with 271k preference pairs (approximately $1.75 \times 10^{19}$ FLOPs). Comparing against DPO with MV additionally allows for an inference FLOPs-adjusted comparison between a model explicitly optimized for alignment and our \rlhfmethod approach. This allows us to test whether \rlhfmethod is not only a better test-time alignment approach, but a better overall alignment approach when significant compute is available at test-time. 

\paragraph{Datasets} We evaluate again on \textbf{GSM8K}; we add \textbf{MATH500}, a verified subset of the MATH  test dataset that contains advanced mathematical problems across algebra, geometry, and calculus \citep{hendrycksmath2021},     \textbf{MMLU-Redux} \citep{gema2025mmlu}, a refined subset of commonsense and academic knowledge questions, \textbf{TruthfulQA}  \citep{truthfulqa}, which contains misleading questions designed to elicit truthful responses, and \textbf{IFEval} \citep{ifeval}, which measures adherence to complex multi-step instructions. Among these datasets, IFEval is an open-ended generation and, therefore, requires the use of MBR with ROUGE-$1$ instead of MV and WMV. For simplicity, we still refer to the method as MV and WMV to denote MBR from base model samples and the weighted version with importance sampling, respectively, in our comparisons.

\paragraph{Sampling Configurations} 

For all datasets, we sampled the model with a temperature of 1.0 and prompted it to provide reasoning. Complete prompts are available in Appendix \ref{appendix:prompts}. As with the mathematical reasoning experiments, we tuned the $\beta$
 parameter (in this case, $\beta = 0.5$), for \textsc{T\"ulu3-8B-RM}, using $128$ samples from the GSM8K training data to achieve an average acceptance rate of $50\%$, and using the same $\beta$ for all datasets. %Our baseline Pass@1 results differ from those presented in the original \textsc{T\"ulu3} paper \citep{tulu3} due to differences in evaluation methodology.  We provide a detailed explanation of these differences and a direct comparison of results in Appendix~\ref{sec:evaldifferences}.

\paragraph{Results}

The results of the general alignment experiments are summarized in Table \ref{table:results}.  Figure \ref{fig:questoutperform} plots the average error rate across all of the datasets as a function of the floating point operations (FLOPS), and Appendix \ref{appendix:allalignmentplots} contains all of the error plots for each individual problem. Similar to our task-specific findings, \rlhfmethod, MV, and WMV consistently reduce error rates across all five general alignment datasets as computation increases, with MV exhibiting early saturation. However, similar to the results from GSM-Symbolic, MV outperforms both WMV and BoN. MV applied to the DPO model begins with lower error rates but saturates earlier, ultimately reaching error rates that are worse than \rlhfmethod's final results. Furthermore, we observe that BoN's inflection point occurs at a lower computational budget than observed in task-specific experiments. This suggests that for real-world alignment problems, BoN is not an effective test-time approach, while \rlhfmethod maintains its improvement trajectory even with general RMs.

\begin{table}
\centering
\setlength{\tabcolsep}{4pt} % 
\begin{tabular}{l|cccccc c}
\toprule
\textbf{Method} & \textbf{MATH500} & \textbf{GSM8K} & \textbf{TQA} & \textbf{MMLU-Redux} & \textbf{IFEval} & \textbf{Avg} \\
& \scriptsize{(0 shot, CoT)} & \scriptsize{(0 shot, CoT)} & \scriptsize{(MC2, 0 shot, CoT)} & \scriptsize{(0 shot, CoT)} & \scriptsize{(prompt loose)} & \scriptsize{} & \\
\midrule
\multicolumn{1}{@{}l}{\textbf{\textsc{\textsc{T\"ulu3-8B-SFT}}}} & \multicolumn{6}{c}{} \\
\textsc{BoN} & 31.6 & 74.3 & 45.3 & 57.1 & 59.3 & 53.5 \\
\textsc{MV} & 49.4 & 86.3 & 43.7 & 59.4 & 72.6 & 62.3 \\
\textsc{WMV} & 53.0 & 85.2 & 46.1 & 62.2 & 49.5 & 59.2 \\
{\rlhfmethod} & \textbf{60.4} & \textbf{88.2} & \textbf{48.5} & 62.2 & \textbf{77.6} & \textbf{67.4} \\ 
\midrule
\multicolumn{1}{@{}l}{\textbf{\textsc{\textsc{T\"ulu3-8B-DPO}}}} & \multicolumn{6}{c}{} \\
\textsc{MV} & 55.8 & {87.9} & 45.4 & \textbf{62.7} & {76.2} & 65.6 \\
\bottomrule
\end{tabular}
\caption{
\textbf{Overview of the results on general alignment.} Results show accuracy percentages with fixed 1024 sampled solutions across different methods and models. Our proposed \rlhfmethod outperforms other approaches on most benchmarks, achieving the highest scores on all datasets when applied to \textsc{T\"ulu3-8B-SFT}, and attaining an average performance better than \textsc{T\"ulu3-8B-DPO}. The highlighted cells indicate best result per benchmark.}
\label{table:results}
\end{table}

\section{Related Work}

\paragraph{MCMC for Text Generation.} 

Our work builds on \method\citep{quest}, which uses MCMC to sample diverse translations. While \method successfully generates multiple translations, its effectiveness as a selection mechanism for a single high-quality response remained unexplored until now. Earlier works have applied MCMC approaches to both autoregressive and masked language models (MLMs). For instance, \citet{cgmh} and \citet{treesearchcgmh} use MH with a proposal distribution that makes token-level modifications for constrained generation tasks.
% where direct sampling is not straightforward,

In the case of MLMs, previous works have explored various forms of Gibbs sampling \citep{berglund2015bidirectional,su2018incorporating,berthasmouth,yamakoshi-etal-2022-probing}.
However, as \citet{goyal2022exposing} show, the conditional distributions from MLMs result in invalid Gibbs samplers. In response, they propose an MH correction on the masked conditionals, resulting in higher quality generations. Building on this, \citet{mixandmatch} and \citet{forristal2023block} apply MLMs for controlled text generation.
Furthermore, several works  have adapted  Hamiltonian MCMC algorithms originally designed for high-dimensional continuous distributions \citep{hmc, Neal2011ProbabilisticIU} to discrete text generation \citep{kumar2022gradientbased, qin2022cold,amini2023, du2023principled}.

\paragraph{Test-Time Scaling.}

Test-time scaling methods have emerged as an important approach for improving LM performance without additional training. While our work focuses on local approximations to the optimal aligned distribution from Eq.~\ref{eq:modifiedgibbs}, a parallel line of research explores heuristic search strategies. Best-of-$n$ (BoN; \citealp{gao2022scalinglawsrewardmodel,stienon2020learning}), majority voting (MV; \citealp{selfconsistency}), weighted majority voting (WMV; \citealp{li-etal-2023-making}) are examples of such approaches. As we outline in\S~\ref{sec:is}--\ref{sec:bon}, BoN and WMV can be interpreted as doing test-time alignment.

The RMs we consider in this work are outcome-based RMs, they only work on full generations. Several recent methods use process-based RMs (PRMs) that evaluate partial generations  \citep{lightman2023letsverifystepstep,wang2024mathshepherdverifyreinforcellms,uesato2022solvingmathwordproblems}. Many techniques have been developed to take advantage of PRMs through guided beam search \citep{xie2023selfevaluationguidedbeamsearch}, Monte-Carlo tree search (MCTS; \citealp{liu2024dontthrowawayvalue,zhang2023planninglargelanguagemodels}), and REBASE \citep{wu2024inferencescalinglawsempirical}.  However, as outlined in \cite{deepseekai2025deepseekr1}, PRMs face significant practical limitations.%: defining fine-grained reasoning steps is challenging, accurately evaluating intermediate steps is difficult (with automated methods yielding unsatisfactory results and manual annotation failing to scale), and PRMs introduce major reward hacking risks while requiring additional computational resources for retraining.

Furthermore, recent theoretical work has analyzed test-time scaling methods, particularly BoN sampling. \citet{farinhas2025rerankinglawslanguagegeneration,schaeffer2025largelanguagemonkeyspower} derive BoN scaling laws relating the number of generated samples to performance gains. \citet{gui2024bonbonalignmentlargelanguage} prove that BoN is optimal for the trade-off between win-rate against the base model and KL divergence from the base model. \citet{beirami2025theoreticalguaranteesbestofnalignment} provide win-rate guarantees and derive a closed-form probability mass function for the BoN policy, along with a new KL estimator. \citet{yang2024asymptoticslanguagemodelalignment} show that BoN is asymptotically equivalent to the optimal policy for reward maximization under KL constraints, assuming a memoryless LM and linear reward function. Our work extends this research direction by establishing a novel approximation connecting the BoN distribution with $n$ samples to the optimal aligned distribution with parameter $\beta$ (\S~\ref{sec:bon}).
%\gf{add a paragraph on test time scaling laws! underlying properties of pass@k cite reranking laws\citep{farinhas2025rerankinglawslanguagegeneration}.}\gf{monkey scaling laws\citep{brown2024largelanguagemonkeysscaling}  }

%\gf{Our work has established new approximation for the BoN distribution that establishes a connection between the number of samples $n$ and regularizer parameter $\beta$. This sampling method as attracted some recent other theoretical attention \citep{yang2024asymptoticslanguagemodelalignment,beirami2025theoreticalguaranteesbestofnalignment,gui2024bonbonalignmentlargelanguage}.}

%\gf{BoN sampling has been the subject of alot of theoretical analysis. It has also attracted some recent theoretical attention  i.e. BoNBoN Alignment paper \citep{gui2024bonbonalignmentlargelanguage} and their related work section.  they say Bon is optimal instead of reward maximize for some kl budget.}
%\gf{guarantees of win rate \citep{beirami2025theoreticalguaranteesbestofnalignment} show a closed form probability mass function of the BoN policy in discrete case and provide a new KL estimator for it.(the thing one from the first part ) }
%\gf{ultra simplistic assumptions - memory less channel and linear.  the optimality in terms of minimizing the cross entropy given an upper bounded KL, and show that BoN is asymptotically equivalent to the optimal policy, which is in line with our findings.\citep{yang2024asymptoticslanguagemodelalignment}}

\section{Conclusion and Future work}

We introduced \rlhfmethod, a test-time alignment method that can sample from the optimal aligned distribution without any model retraining (requiring only a reward model). Our empirical results consistently demonstrate that \rlhfmethod outperforms both search-based approaches that attempt to maximize imperfect reward models (like BoN) and principled alternatives that rely on independent samples from the base LM (such as WMV and MV). 
Additionally, \rlhfmethod outperforms DPO-tuned models even when they are allowed to match \rlhfmethod's compute budget at test-time.

By enabling strong alignment results without model retraining, \rlhfmethod opens new possibilities for deploying and improving language models in resource-constrained environments and enabling the use of private RMs with closed-source LMs.

Looking ahead, we believe this work sets the stage for further exploration of test-time alignment methods. While we have demonstrated success across several benchmarks, several important directions remain unexplored. These include the development of more sophisticated proposal distributions and using MCMC as a tool to understand the meta-generation strategies that emerge in RL-trained reasoning models. 

%\section*{Acknowledgments}
%We thank Alisa Liu, André Martins, Guilherme Pires, Hamish Ivison and Assaf Harel for their helpful and constructive feedback on the initial versions of the paper. This work was supported in part by NSF grant 211353 and NVIDIA resources provided through the National AI Research Resource Pilot (NAIRR).

\bibliography{custom}
\bibliographystyle{iclr2026_conference}
\appendix  % Switches to appendix mode

\newpage 
\section{Best-of-$n$ Sampling as Test-Time Alignment}
\label{appendix:bonproof}

Best-of-$n$ (BoN) sampling is a widely used technique for improving model performance by selecting the highest scoring sample among $n$ candidates. In this section, we show that BoN sampling can be interpreted as an implicit form of test-time alignment that approximates the \textbf{mode} of the target distribution $\pi(y\mid x)$ in Eq.~\ref{eq:modifiedgibbs}. In \S\ref{sec:maxpdf}, we derive the probability density function (pdf) for BoN sampling. Next, in \S\ref{sec:evt}, we use extreme value theory to obtain a tractable approximation of rewards of this distribution, and in \S\ref{sec:mixedgauss} derive an expression for the distribution of rewards of the target density $\pi(y \mid x )$. Finally, in \S\ref{sec:laplaceapprox}, employing a Laplace approximation of the reward distribution, we establish an explicit relationship between $n$ and the effective parameter $\beta$ of the aligned distribution $\pi(y\mid x)$.

\subsection{Distribution of Maximum Reward}
\label{sec:maxpdf}

In BoN sampling, given a prompt $x$, we generate $n$ candidate responses $y_1,y_2, \ldots,y_n$ independently from the base LM $p_\text{LM}(y \mid x)$. Each response is evaluated using the RM $r(y,x)$, and the candidate with the highest reward is chosen as the final output:
\begin{equation}
y^\ast(n) = \arg\max_{y_i} r(x,y_i).
\end{equation}
Let us assume that the reward function $r(x,y)$ is (locally) injective, that is,  each $y$ maps to a unique reward value. 
Then, if we denote by $p(r \mid x)$ the probability density (pdf) of reward values for a particular $x$ under the base LM $p_\text{LM}(y \mid x)$ a change of variables yields:
\begin{equation}
p(r\mid x)=p_\text{LM}\Big(y=r^{-1}(r,x)\mid x\Big)\Big| J_{r^{-1}}(r)\Big| ,
\end{equation} where $r^{-1}$ is the inverse mapping and $|J_{r^{-1}}(r)|$ is its Jacobian determinant.

Given $n$ independent samples, the cumulative distribution function (cdf) $F_R(r|x)$ we can write the cdf for the maximum reward $r_{\text{max}}$ as
\begin{equation}
F_{\text{bon}}(r^{(n)}_{\text{max}}|x) = F_R(r^{(n)}_{\text{max}} \mid x)^{n},\label{eq:truerewardist}\end{equation} and the respective pdf as 
\begin{equation}
p_{\text{bon}}(r^{(n)}_{\text{max}}|x) = n\, F_R(r^{(n)}_{\text{max}} \mid x)^{n-1}\, p(r=r^{(n)}_{\text{max}} \mid x).\label{eq:truerewardist}\end{equation} 
Since each candidate $y$ corresponds to a reward $r(x,y)$, the probability density over the best-of-$n$ response can be written as:
\begin{equation}
p(y^\ast(n)|x) = n\, F_R\big(r(x,y^\ast(n)) \mid x\big)^{n-1}\, p_{\text{LM}}(y^\ast(n)|x).\label{eq:boncorrect}
\end{equation}

Eq.~\ref{eq:boncorrect} shows how the BoN procedure emphasizes the upper tail of the reward distribution. The term $n F_R(r \mid x)^{n-1}$ becomes increasingly significant as $n$ grows, effectively pushing the probability mass towards reward values for which $F_R(r \mid x) \approx 1$.

\subsection{Extreme Value Theory}
\label{sec:evt}

Recall that our goal was to establish an explicit relationship between the number of samples we maximize over $n$ and the parameter $\beta$ of the aligned distribution in Eq. \ref{eq:modifiedgibbs}. One way to establish this relationship is through the reward distribution. However, even assuming $p(r \mid x)$ is normal, the mode of the distribution in Eq.~\ref{eq:truerewardist} lacks a closed-form expression.

Following classic results from extreme value theory \citep{Fisher_Tippett_1928,David1960StatisticsOE}, we know that the distribution of the maximum (appropriately rescaled) converges to a Gumbel distribution. For the specific case where we assume that for a given prompt $x$ the reward function values are distributed as a mixture of two Normal distributions with means $\mu_1(x)$ and $\mu_2(x)$, variances $\sigma_1^2(x)$ and $\sigma_2^2(x)$, and mixture weights $w_1(x)$ and $w_2(x)$ (where $w_1(x) + w_2(x) = 1$). Appendix~\ref{eq:empiricaldist} provides empirical evidence that this assumption is reasonable. The tail behavior is dominated by the Gaussian component with the heavier tail, typically the one with larger variance (or larger mean if variances are equal). 
\begin{equation}
    1 - F_{r_{\max}^{(n)} }(r) \approx w_d(x) \left( 1 - \Phi \left( \frac{r-\mu_d(x)}{\sigma_d(x)}\right) \right),
\end{equation} where $\Phi$ is the standard normal CDF, $d$ the index of the component with the largest variance, i.e., $d = \arg \max_i \sigma_i^2(x)$, or if variances are equal, $d = \arg \max_i \mu_i(x)$ is the index of the component with the largest mean.

\begin{figure}[htb]
    \centering
    \includegraphics[width=0.5\linewidth]{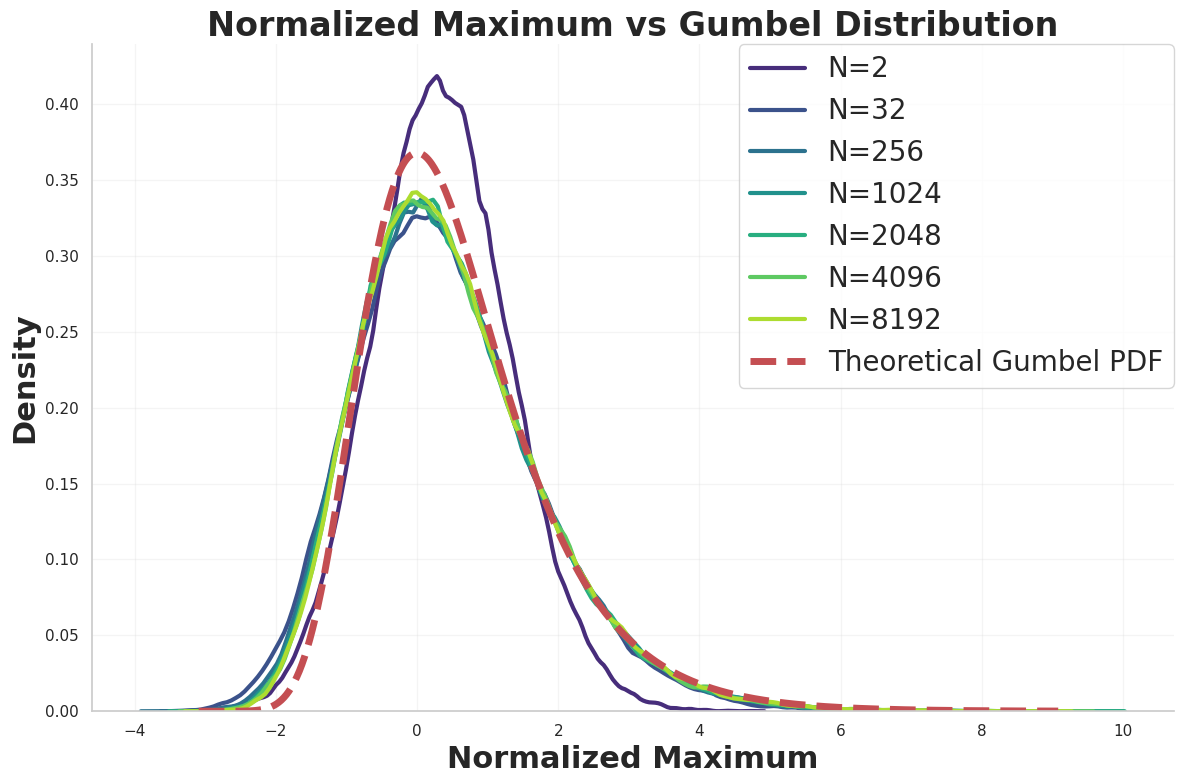}
    \caption{Distribution of the normalized maximum reward $({r^{(n)}_{\text{max}} - a_n})/{b_n}$ for varying $n$, overlaid with the standard Gumbel distribution. The empirical distribution is estimated using 10,000 trials, each consisting of $n$ random samples drawn from a Normal distribution. The fit between the empirical distribution and the Normal distribution improves as $n$ increases, showing good agreement for \(n \geq 32\).}
    %\caption{ Distribution of normalized maximum reward ($(r^{n}_\text{max}-a_n)/b_n$) for varying $n$, overlaid with the Gumbel approximation. The fit improves as $n$ increases, with good agreement for $n \geq 32$. The empirical is a kde from from 10000 trials of $n$ random gaussian samples. }
    \label{fig:gumbeldist}
\end{figure}

The work of \citet{1beb852f-e02d-3c6a-88a8-d80908be5528} provides asymptotic expressions for location and scale parameters. In particular, one finds that the maximum reward is approximately:
\begin{equation}
r_{\max}^{(n)} \approx a_n \equiv \mu_d(x) + \sigma_d(x) \left(\sqrt{2\log n_d} - \frac{(\log\log n_d + \log(4\pi))}{2\sqrt{2\log n_d}} \right),
\end{equation} and fluctuations about this location are on the order of $b_n \approx \frac{\sigma_d(x)}{\sqrt{2\log n_d}}$, and where $n_d = n w_d(x)$ denotes the effective sample count from the dominant normal component. So in the limit of $n\rightarrow \infty$ the distribution of maximum reward can be expressed as:
\begin{equation}
p(r_{\max}^{(n)}|x) \approx \text{Gumbel}(a_n,b_n).
\end{equation} 
Figure~\ref{fig:gumbeldist} compares this Gumbel approximation to the empirical distribution of maximum rewards, showing a close fit for $n\geq32$. 

\subsection{Distribution of Rewards of $\pi( y \mid x)$}
\label{sec:mixedgauss}
Given the aligned distribution $\pi(y\mid x)$ from Eq.~\ref{eq:modifiedgibbs}, we want an expression for the distribution of rewards under samples from $\pi$. 
Following the empirical observations from Appendix~\ref{eq:empiricaldist}, let us assume that for a given prompt $x$, the reward values under the base language model $p_\text{LM}(y \mid x)$ follow a mixture of two Normal distributions with means $\mu_1(x)$ and $\mu_2(x)$, variances $\sigma_1^2(x)$ and $\sigma_2^2(x)$, and mixture weights $w_1(x)$ and $w_2(x)$ (where $w_1(x) + w_2(x) = 1$), i.e.:
\begin{equation*}
p(r \mid x) = w_1(x) \frac{1}{\sqrt{2\pi\sigma_1^2(x)}}\exp\left(-\frac{(r-\mu_1(x))^2}{2\sigma_1^2(x)}\right) + w_2(x) \frac{1}{\sqrt{2\pi\sigma_2^2(x)}}\exp\left(-\frac{(r-\mu_2(x))^2}{2\sigma_2^2(x)}\right).
\end{equation*}
Under $\pi(y \mid x)$, the distribution of rewards can be derived through a change of variables. The probability density of rewards under the policy, denoted as $\pi(r \mid x)$, is proportional to:
\begin{equation}
\pi(r \mid x) \propto \exp\left(\frac{r}{\beta}\right)p(r \mid x).
\end{equation}
Substituting the mixture form of $p(r \mid x)$ and combining terms in the exponent and through completion of the square in the exponents for each component, we can show that this is equivalent to:
\begin{align}
\pi(r \mid x) &\propto w_1(x) C_1 \frac{1}{\sqrt{2\pi\sigma_1^2(x)}}\exp\left(-\frac{1}{2\sigma_1^2(x)}\left(r-\mu_1(x)-\frac{\sigma_1^2(x)}{\beta}\right)^2\right) \\
&+ w_2(x) C_2 \frac{1}{\sqrt{2\pi\sigma_2^2(x)}}\exp\left(-\frac{1}{2\sigma_2^2(x)}\left(r-\mu_2(x)-\frac{\sigma_2^2(x)}{\beta}\right)^2\right),
\end{align}
where $C_1 = \exp\left(\frac{\mu_1(x)}{\beta} + \frac{\sigma_1^2(x)}{2\beta^2}\right)$ and $C_2 = \exp\left(\frac{\mu_2(x)}{\beta} + \frac{\sigma_2^2(x)}{2\beta^2}\right)$. 

Therefore, the aligned distribution of rewards remains a mixture of Normals with adjusted mixture weights and  means, and preserved variances:
\begin{equation}
\pi(r \mid x) = w_{\pi,1} \mathcal{N}(r; \mu_{\pi,1}(x,\beta), \sigma_1^2(x)) + w_{\pi,2} \mathcal{N}(r; \mu_{\pi,2}(x,\beta), \sigma_2^2(x)), \label{eq:alignedrewardist}
\end{equation}
where $\mu_{\pi,i}(x,\beta) = \mu_i(x) + {\sigma_i^2(x)}/{\beta}$ for $i \in \{1,2\}$, and the adjusted mixture weights are given by:
\begin{equation}
w_{\pi,1} = \frac{w_1(x) C_1}{w_1(x) C_1 + w_2(x) C_2}, \quad w_{\pi,2} = \frac{w_2(x) C_2}{w_1(x) C_1 + w_2(x) C_2}.
\end{equation}

This result shows that $\beta$ shifts the mean of each component by a factor proportional to that component's variance and inversely proportional to the parameter $\beta$, while preserving the variances of the original reward distributions. Additionally, as $\beta$ approaches zero, the mixture weight of the component with the larger variance (or larger mean if variances are equal) approaches 1, \textbf{causing the aligned distribution to collapse to a single Gaussian with an increasingly high mean}.

For this reason, we write the reward target density expression as approximately only the dominant Gaussian:
\begin{equation}
\pi(r \mid x) \approx \mathcal{N}(r; \mu_{\pi,d}(x,\beta), \sigma_d^2(x)), \label{eq:dominantgaussian}
\end{equation}
where $d = \arg \max_i \sigma_i^2(x)$ is the index of the component with the largest variance, or if variances are equal, $d = \arg \max_i \mu_i(x)$ is the index of the component with the largest mean.

\subsection{Relating $n$ and $\beta$ via Mode Matching}
\label{sec:laplaceapprox}

The Gumbel approximation has a mode that can be explicitly expressed as its location parameter $a_n$, and the reward distribution of the aligned distribution in Eq.~\ref{eq:dominantgaussian} as $\mu_{\pi,d}(x,\beta)$. 

Following the results from Appendix~\ref{sec:mixedgauss}, the distribution of rewards under the target density $\pi(r|x)$ is also normal, but with mean $\mu_{\pi,d}(x,\beta) = \mu_d(x) + {\sigma_d(x)^2}/{\beta}$.

By matching this mode with the location parameter $a_n$ of the Gumbel distribution for BoN sampling i.e.:
\begin{equation}
\mu_d(x) + \frac{\sigma_d(x)^2}{\beta}  = \mu_d(x) + \sigma_d(x) \left(\sqrt{2\log n_d} - \frac{(\log\log n_d + \log(4\pi))}{2\sqrt{2\log n_d}} \right),
\end{equation} we obtain:
 \begin{equation}
    \beta^\ast = \frac{\sigma_d(x)}{\sqrt{2\log n_d} - \frac{(\log\log n_d + \log(4\pi))}{2\sqrt{2\log n_d}}}.
\end{equation}

\begin{figure}[htb]
    \centering
    \includegraphics[width=0.5\linewidth]{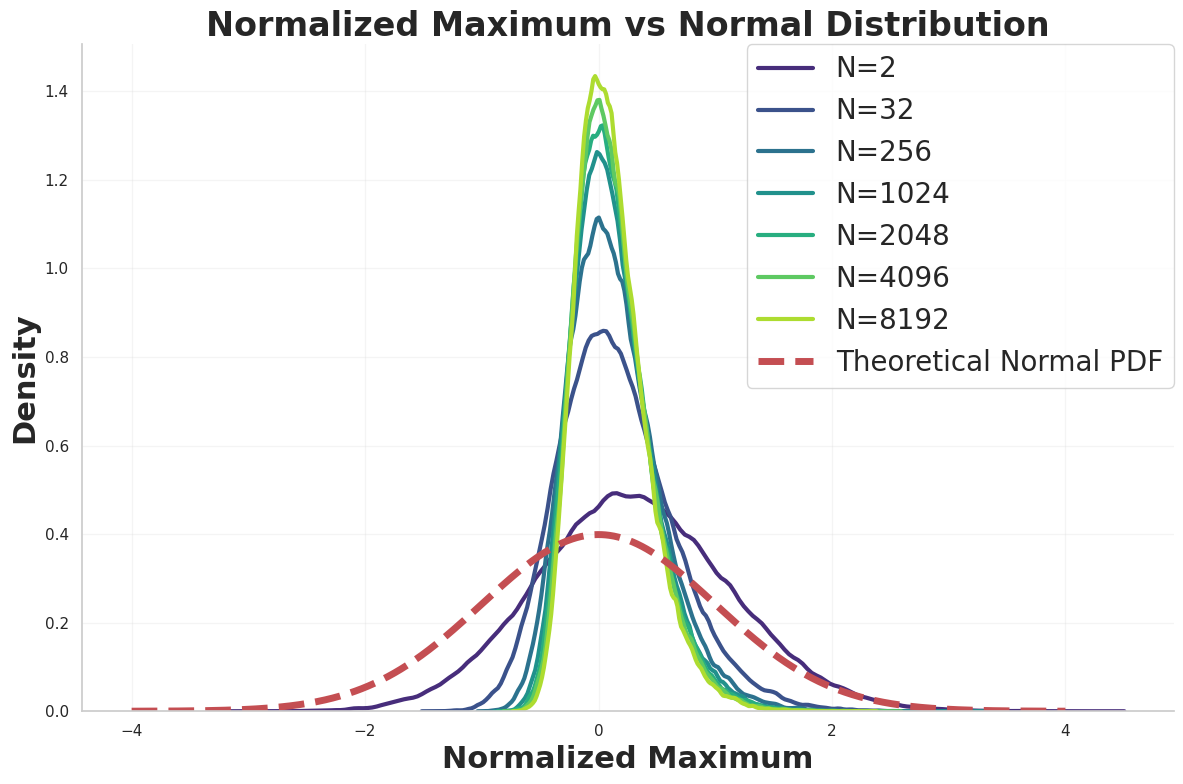}
    \caption{ Distribution of the normalized maximum reward $({r^{(n)}_{\text{max}} - \mu_{\pi,d}(x,\beta^\ast) })/\sigma_d(x)$ for varying $n$, overlaid with the standard Normal distribution. The empirical distribution is estimated using 10,000 trials, each consisting of $n$ random samples drawn from a Normal distribution. While the mode of our approximation matches, the approximation does not capture the variance of the empirical distribution.}
    \label{fig:normaldist}
\end{figure}

Figure~\ref{fig:normaldist} compares the empirical reward distribution to the target distribution $\pi(r|x)$ when $\beta = \beta^\ast$. The approximation closely matches the mode of the empirical distribution of maximum rewards, though it struggles to capture the variance accurately. The variance of the BoN reward distribution decreases as a function $n$, while the one from $\pi( y \mid x)$ stays constant.  

\newpage 
\section{RLHF as Bayesian Inference}
\label{appendix:rlhfvi}

For a fixed prompt $x$, the RLHF objective from Eq.~\ref{eq:rlhfobj}, i.e.:

\begin{equation*}
\mathcal{L}(\theta, x)
=  \mathbb{E}_{x\sim\mathcal{D}} \big[ \mathbb{E}_{y \sim q_\theta(y\mid x)}\left[ r_\phi(y,x)/\beta \right] - D_{\text{KL}} \left( q_\theta(y\mid x)\ \|\ p_{\text{LM}}(y\mid x) \right) \big],
\end{equation*}

where $q_\theta(y\mid x)$ is the learned policy, is maximized by the posterior
\begin{equation*}
p(y \mid \gamma=1,x)
= \frac{1}{Z_\beta(x)}\, p_\text{LM}(y \mid x)\exp \left({r(y, x)}/{\beta} \right),
\end{equation*}
which we denote by $\pi_\beta(y\mid x)$.

\medskip
\noindent\emph{Proof.}

Following the definitions in \S~
\ref{sec:background}, $\gamma=1$ denotes a “preferred’’ response under the human preference model, and $Z_\beta(x)$ is the normalizing constant. To prove this, we follow the standard variational inference derivation starting from the log evidence $\log p(\gamma=1\mid x)$.

We begin by writing
\begin{equation*}
\log p(\gamma=1\mid x)
=  \mathbb{E}_{y \sim q_\theta(y\mid x)} \left[ \log p(\gamma=1\mid x) \right].
\end{equation*}
By Bayes' rule we can rewrite the log evidence as
\begin{equation*}
\log p(\gamma=1\mid x)
=  \mathbb{E}_{y \sim q_\theta(y\mid x)}
   \left[ \log  \frac{p(y,\gamma=1\mid x)}{p(y\mid \gamma=1,x)} \right].
\end{equation*}

We now multiply and divide inside the logarithm by $q_\theta(y\mid x)$:
\[
\log p(\gamma=1\mid x)
=  \mathbb{E}_{y \sim q_\theta(y\mid x)}
   \left[ \log  \frac{p(y,\gamma=1\mid x)\, q_\theta(y\mid x)}
                  {p(y\mid \gamma=1,x)\, q_\theta(y\mid x)} \right].
\]

This yields the standard variational decomposition:
\[
\log p(\gamma=1\mid x)
=  \mathbb{E}_{y \sim q_\theta(y\mid x)}
     \left[ \log  \frac{p(y,\gamma=1\mid x)}{q_\theta(y\mid x)} \right]
  + D_{\mathrm{KL}}\!\big(q_\theta(y\mid x)\ \|\ p(y\mid \gamma=1,x)\big).
\]

Next we expand the joint distribution $p(y,\gamma=1\mid x) = p(\gamma=1\mid y,x)\, p_\text{LM}(y\mid x)$ which decomposes further into:
\[
\log p(\gamma=1\mid x)
= \mathbb{E}_{y \sim q_\theta(y\mid x)}\!\left[ \log p(\gamma=1\mid y,x) \right]
  - D_{\mathrm{KL}}\!\big(q_\theta(y\mid x)\ \|\ p_\text{LM}(y\mid x)\big)
  + D_{\mathrm{KL}}\!\big(q_\theta(y\mid x)\ \|\ p(y\mid \gamma=1,x)\big).
\]

We now plug in the preference model (\S~\ref{sec:background})
\[
p(\gamma=1\mid y,x)
= \exp\!\left( (r(y,x) - \max_y r(y,x))/\beta \right),
\]
and note that both $\max_{y} r(y,x)$ and $\log p(\gamma=1\mid x)$ are constants that depend only on $x$. Thus we can rewrite the previous identity as
\[
c
= \mathbb{E}_{y \sim q_\theta(y\mid x)}\!\left[ \frac{r(y, x)}{\beta} \right]
  - D_{\mathrm{KL}}\!\big(q_\theta(y\mid x)\ \|\ p_\text{LM}(y\mid x)\big)
  + D_{\mathrm{KL}}\!\big(q_\theta(y\mid x)\ \|\ p(y\mid \gamma=1,x)\big),
\]
for some constant $c$ depending only on $x$.

Rearranging gives
\[
c
- \mathbb{E}_{y \sim q_\theta(y\mid x)}\!\left[ \frac{r(y, x)}{\beta} \right]
+ D_{\mathrm{KL}}\!\big(q_\theta(y\mid x)\ \|\ p_\text{LM}(y\mid x)\big)
= D_{\mathrm{KL}}\!\big(q_\theta(y\mid x)\ \|\ p(y\mid \gamma=1,x)\big).
\]

Since the left-hand side is exactly $c - \mathcal{L}(\theta,x)$, minimizing the KL divergence
\[
D_{\mathrm{KL}}\big(q_\theta(y\mid x)\ \|\ p(y\mid \gamma=1,x)\big)
\]
is equivalent to maximizing the RLHF objective $\mathcal{L}(\theta,x)$. Therefore the maximizer in $q_\theta$ is
\[
q_\theta(y\mid x) = p(y\mid \gamma=1,x),
\]
i.e., the posterior distribution
\[
p(y \mid \gamma=1,x)
= \frac{1}{Z_\beta(x)}\, p_\text{LM}(y \mid x)\exp \left({r(y, x)}/{\beta} \right).
\]

This shows that $\pi_\beta(y\mid x)$ is the optimal policy for a single prompt.

\section{\rlhfmethod Convergence proof}
\label{sec:convergenceproof}

Let $\pi_\beta(y\mid x)$ be the target distribution, and consider the Metropolis–Hastings chain used by \rlhfmethod with target $\pi_\beta$ and the \method proposal kernel $q(y \mid y^t,x)$. Then the distribution of $y^t$ converges in distribution to $\pi_\beta(y\mid x)$ as $t \to \infty$.

\medskip
\noindent\emph{Proof.}

 The theory comes from the standard Metropolis--Hastings (MH; \cite{mh}) together with the properties of the proposal distribution introduced in \method \citep{quest}. Once we specify (i) the target distribution $\pi_\beta(y\mid x)$  and (ii) a valid proposal kernel, the MH theorem ensures that the resulting Markov chain converges to $\pi_\beta$ in the limit.

The central requirement is that the MH transition kernel satisfies \emph{detailed balance} with respect to~$\pi_\beta$:
\[
\pi_\beta(y^t\mid x)\, T(y^t \rightarrow y)
= 
\pi_\beta(y\mid x)\, T(y \rightarrow y^t),
\]
where $T(\cdot\to\cdot)$ denotes the one-step transition probability of the chain. For the standard MH acceptance rule,
\[
\alpha_\beta(y,y^t)
= \min\!\left(1, \frac{\pi_\beta(y\mid x)\, q(y^t\mid y,x)}
                       {\pi_\beta(y^t\mid x)\, q(y\mid y^t,x)} \right),
\]
\cite{mh} shows that detailed balance holds automatically for any proposal $q(y\mid y^t,x)$ that is \textit{irreducible} and \textit{aperiodic}.

The work of \method \citep{quest} proves that the proposal distribution we use, defined using LLM-based random suffix resampling, is irreducible and aperiodic. An irreducible Markov chain is one that for any two responses $y, y'$ in the support of the LM, the proposal kernel $q(y' \mid y, x)$ has non-zero probability of reaching $y'$ from $y$ in a finite number of proposal steps. In our case, this is ensured by the fact that the proposal mechanism can always select index~0 and regenerate an entire response from scratch.

An aperiodic Markov chain is one in which every state is aperiodic (equivalently, every state has period 1). A state is periodic if returns to that state can occur only after numbers of steps that are all multiples of some integer greater than 1, for example only after an even number of steps, or only after multiples of 3. Otherwise, the state is aperiodic, meaning there is no such mandatory multiple. A simple sufficient condition for aperiodicity is that each state has a non-zero probability of returning to itself in one transition (a self-loop). In our setting, this holds because with positive probability we sample index~0 (which occurs with probability $1/|y|$) and regenerate the entire response, and this regeneration can reproduce the current response with non-zero probability, giving a one-step self-return.

Thus, combining detailed balance with irreducibility and aperiodicity, the Markov chain ergodic theorem\citep{Neal2011ProbabilisticIU} guarantees that the chain convergence to the optimal aligned distribution $\pi_\beta$ in the limit $t\rightarrow \infty$.

\section{Computational Cost of \rlhfmethod}
\label{sec:computation}

As outlined in \cite{quest}, when using the suffix proposal distribution from \method, each step samples, on average, an index at the midpoint of the sentence from the uniform distribution. Assuming a fixed sentence length $N$ for simplicity, this requires generating only $\frac{N}{2}$ new tokens on average per step. With typical transformer-based models, this allows us to reuse the majority of the key-value cache from previous iterations in both the base model and reward model. The reuse extends even beyond half of the computation due to the static prompt. However, unlike when generating independent samples, the computation for each prompt is sequential. We need to first generate the sentence $y^{t}$ before generating the sentence $y^{t+1}$. This means that regardless of the compute capability, compared with sampling independent samples, there is always an inherent latency overhead. %\gf{note that a model o1 reasoning produces a linear reasoning chain so they have the same issue.} 

In summary, for a chain of $T$ steps, \rlhfmethod is expected to generate $\frac{(T+1)N}{2}$ tokens in total: $N$ tokens for the initial hypothesis, plus an average of $\frac{N}{2}$ tokens for each of the remaining $T-1$ steps. In comparison, generating $T$ independent samples requires decoding $T \times N$ tokens. Therefore, for an equal number of samples, \method requires approximately half as many tokens from generation alone than sampling independently, translating to roughly \textbf{half the FLOPs}.

\section{Empirical Observations on Distribution of Rewards}
\label{eq:empiricaldist}
\begin{figure}[htb]
    \centering
    \includegraphics[width=0.6\linewidth]{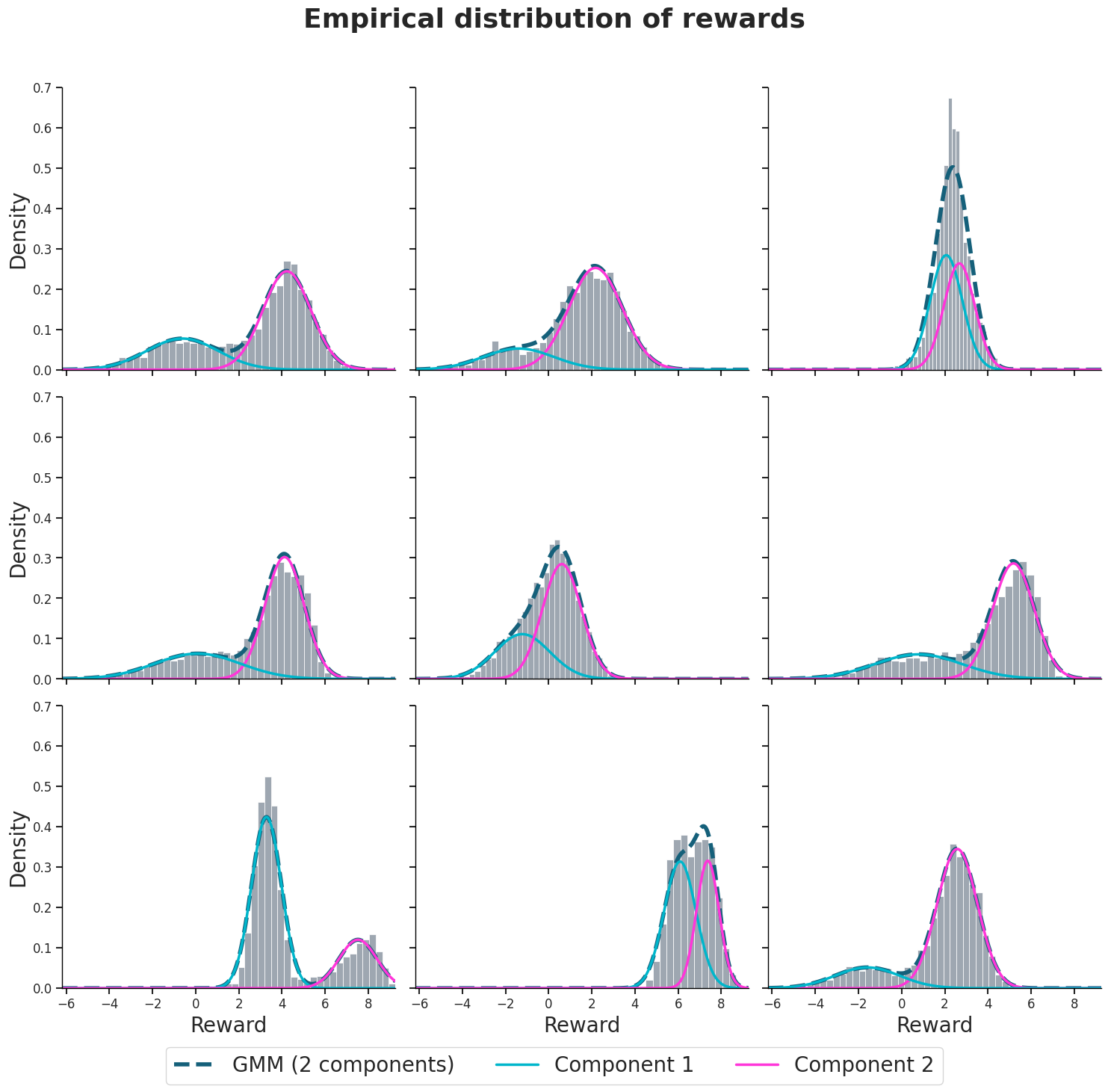}
    \caption{Histogram of rewards assigned by \textsc{T\"ulu3-8B-RM} to $1,024$ responses generated by \textsc{T\"ulu3-8B-SFT} for $9$ randomly sampled prompts from GSM8K. For each prompt, we fit a two-component Gaussian mixture model to characterize the reward distribution. }
    \label{fig:mixtureempirical}
\end{figure}

When analyzing reward model predictions across independently generated responses from the base model for individual prompts, we consistently observed a bimodal distribution forming a two-component Normal mixture with distinct means and variances (Figure \ref{fig:mixtureempirical}). This pattern appeared across the reward models used.

We suspect this bimodal structure directly relates to the Bradley-Terry training objective, which naturally tries to separate the responses into two clusters (preferred vs.~non-preferred). The clusters appear to have a normal distribution most likely because of the central limit theorem, i.e., the predictions result from a neural net that is the sum of billions of random variables.

\section{Prompts Used for Evaluation}  
\label{appendix:prompts}

This appendix documents the prompt templates used across different datasets in our experiments. The prompt for GSM8K is only used in the general alignment experiments. The placeholders (text within \texttt{<\{...\}>}) are dynamically replaced with specific content from each dataset during the experiments.

\subsection{GSM8K Dataset}
\begin{tcolorbox}[colback=gray!5, colframe=gray!40, boxrule=0.5pt]
\begin{minipage}{\linewidth}
\ttfamily
Solve the following grade school math problem step-by-step: <\{question\}>
\end{minipage}
\end{tcolorbox}

\subsection{MATH-500 Dataset}
\begin{tcolorbox}[colback=gray!5, colframe=gray!40, boxrule=0.5pt]
\begin{minipage}{\linewidth}
\ttfamily
Solve the following math problem step-by-step: <\{question\}>\\
Present the answer in LaTex format: \textbackslash boxed\{Your answer\}
\end{minipage}
\end{tcolorbox}

\subsection{Multiple Choice Datasets (TQA, MMLU)}
\begin{tcolorbox}[colback=gray!5, colframe=gray!40, boxrule=0.5pt]
\begin{minipage}{\linewidth}
\ttfamily
Choose the correct answer to the following multiple-choice question about <\{subject\}>.\\
Question: <\{question\}>\\
A). <\{choice\_A\}>\\
B). <\{choice\_B\}>\\
C). <\{choice\_C\}>\\
D). <\{choice\_D\}>\\
Provide your reasoning about the answer and finish your answer with the letter corresponding to the correct option (e.g., A, B, C, or D).
\end{minipage}
\end{tcolorbox}

\newpage
\section{Full General Alignment Plots} 
\label{appendix:allalignmentplots}

\begin{figure}[htb]
    \centering
    \begin{subfigure}[b]{0.45\textwidth}
        \centering
        \includegraphics[width=\textwidth]{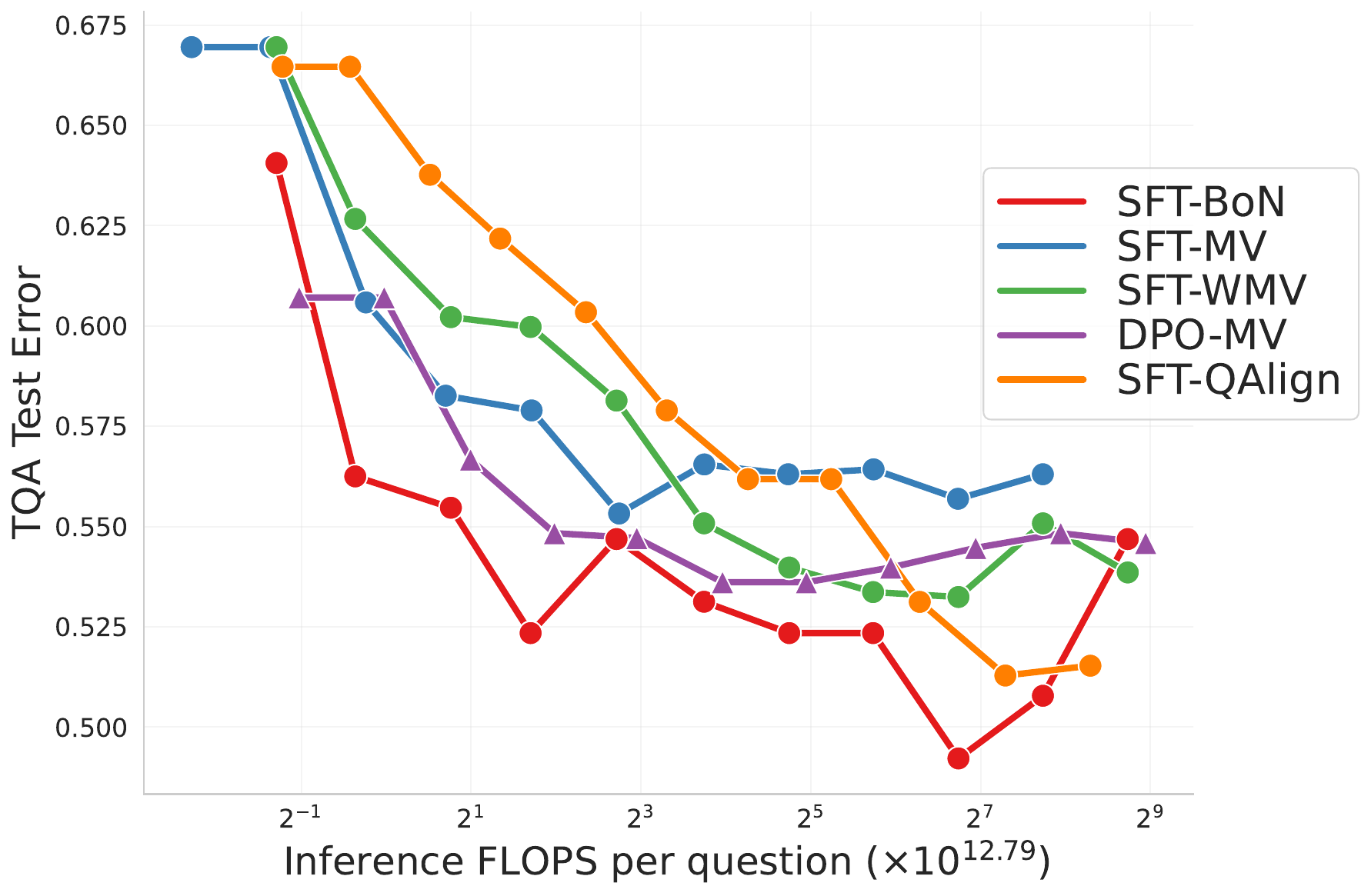}
        %\caption{TQA Tulu alignment}
        %\label{fig:tqa_tulu}
    \end{subfigure}
    \hfill
    \begin{subfigure}[b]{0.45\textwidth}
        \centering
        \includegraphics[width=\textwidth]{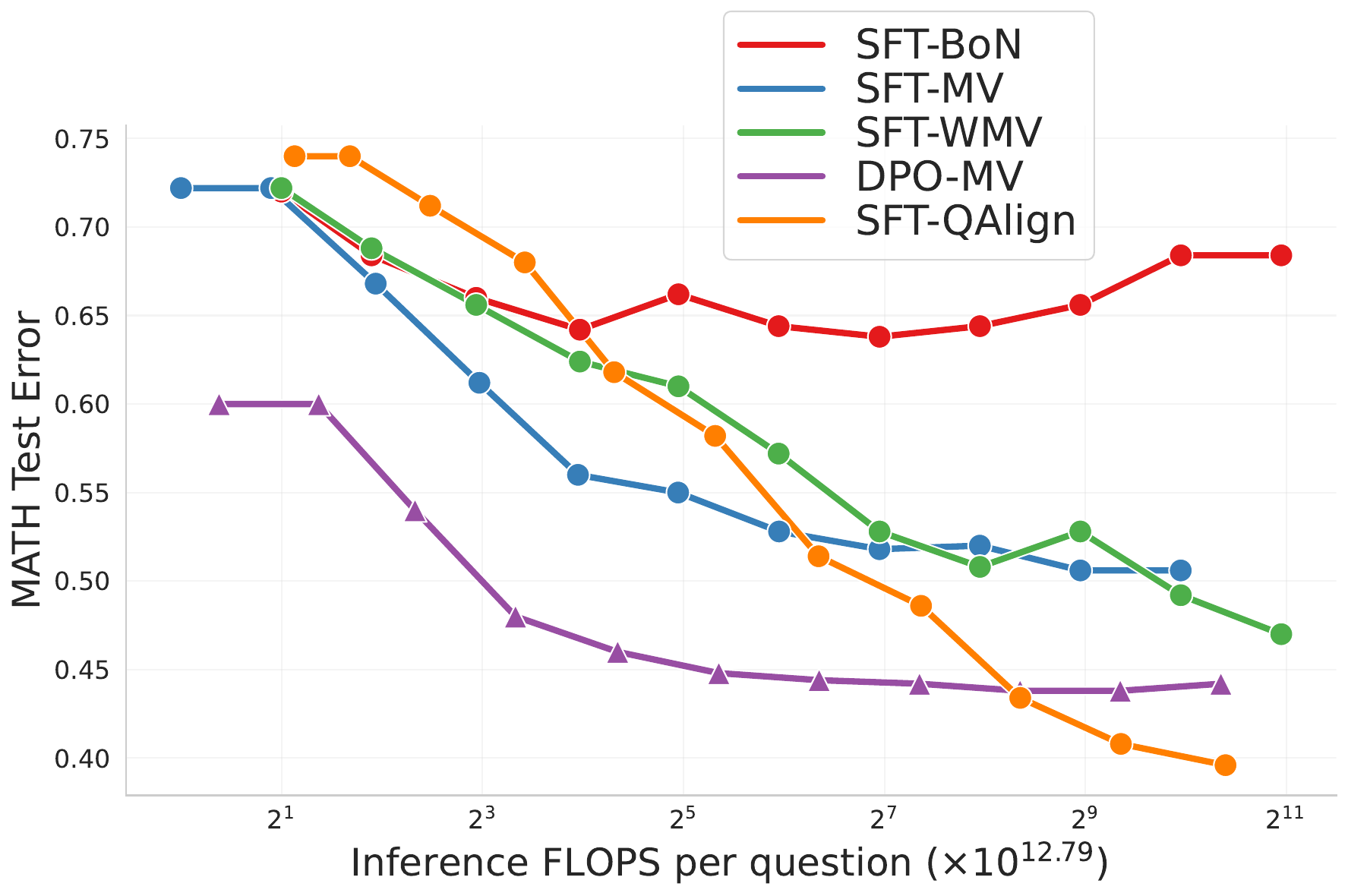}
        %\caption{Math Tulu alignment}
        %\label{fig:math_tulu}
    \end{subfigure}
    
    \vspace{0.5cm}
    
    \begin{subfigure}[b]{0.45\textwidth}
        \centering
        \includegraphics[width=\textwidth]{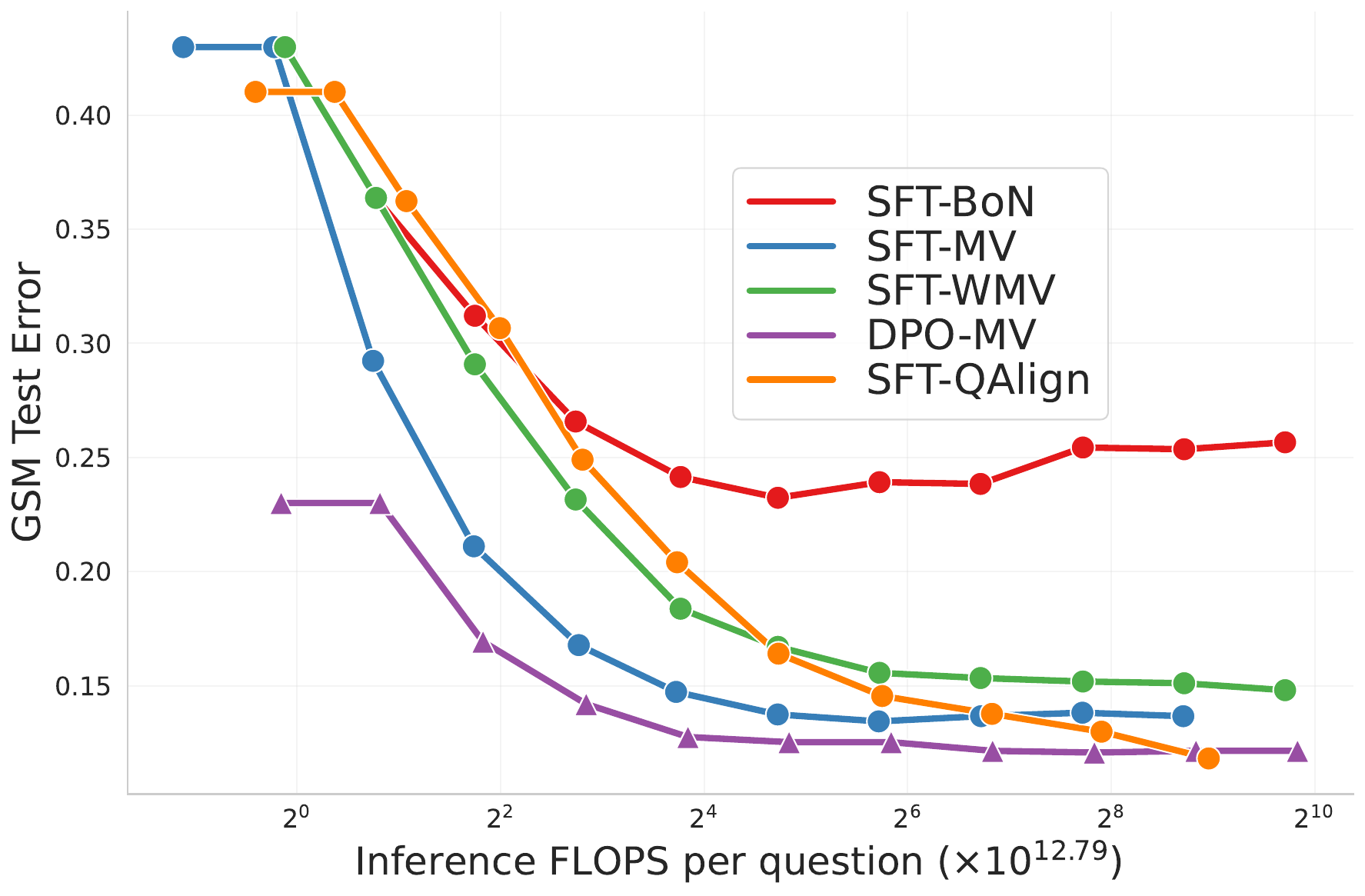}
        %\caption{GSM Tulu alignment}
        %\label{fig:gsm_tulu}
    \end{subfigure}
    \hfill
    \begin{subfigure}[b]{0.45\textwidth}
        \centering
        \includegraphics[width=\textwidth]{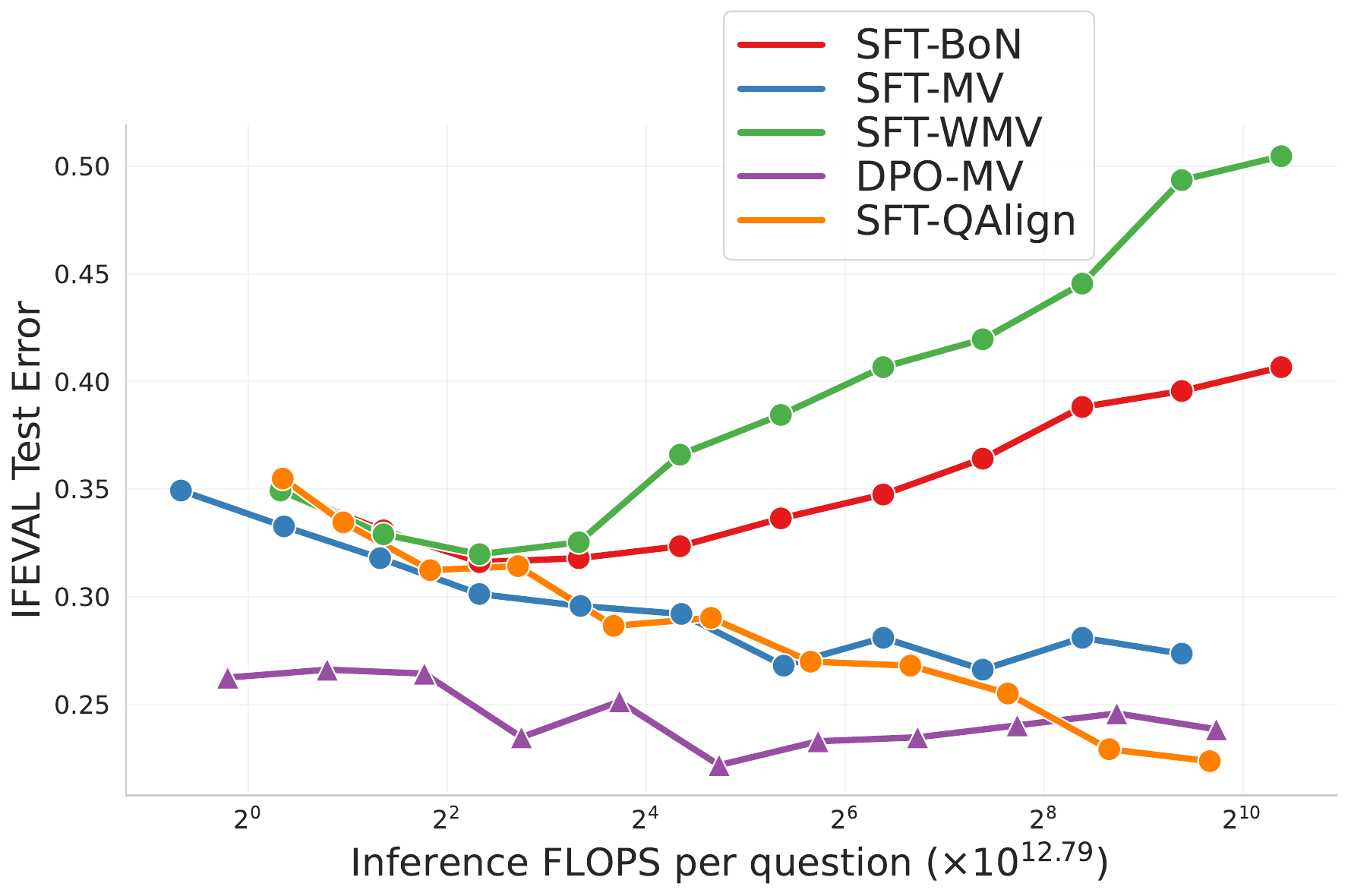}
        %\caption{IFEval Tulu alignment}
        %\label{fig:ifeval_tulu}
    \end{subfigure}
    
    \vspace{0.5cm}
    
    \begin{subfigure}[b]{0.45\textwidth}
        \centering
        \includegraphics[width=\textwidth]{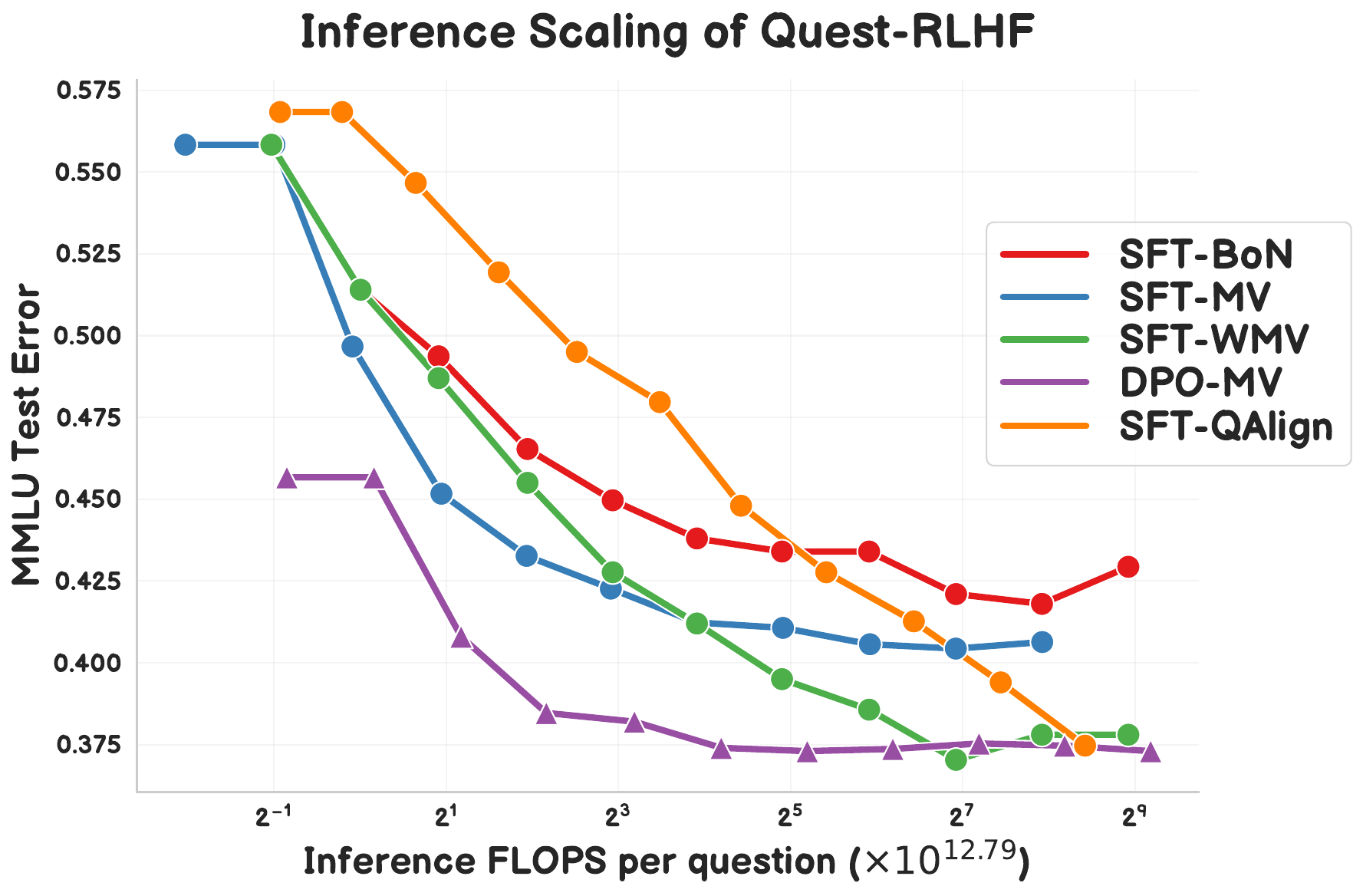}
        %\caption{MMLU Tulu alignment}
        %\label{fig:mmlu_tulu}
    \end{subfigure}
    
    \caption{Error rate across multiple evaluation datasets (GSM8K, MATH500, MMLU-Redux, TruthfulQA, and IFEval) as a function of the floating point operations (FLOPS) in log scale. 
    We compare \textbf{\textcolor{SFT-QUEST}{\raisebox{-0.1em}{\scalebox{1.7}{$\bullet$}}\rlhfmethod with \textsc{T\"ulu3-8B-SFT}}}  against four baselines: \textbf{\textcolor{DPO-MV}{\raisebox{0.0em}{\scalebox{1.2}{$\blacktriangle$}} majority vote (MV) \textsc{T\"ulu3-8B-DPO}}}, and applied to \textsc{T\"ulu3-8B-SFT} the methods \textbf{\textcolor{SFT-BoN}{\raisebox{-0.1em}{\scalebox{1.7}{$\bullet$}} best-of-$n$ (BoN)}},  \textbf{\textcolor{SFT-MV}{\raisebox{-0.1em}{\scalebox{1.7}{$\bullet$}} MV}}, and \textbf{\textcolor{SFT-WMV}{\raisebox{-0.1em}{\scalebox{1.7}{$\bullet$}} weighted MV (WMV)} }. All experiments use temperature 1.0 with reasoning included in model outputs. Note that \textsc{T\"ulu3-8B-DPO} model is the result of doing preference finetuning on the \textsc{T\"ulu3-8B-SFT} with 271k preference pairs. The costs associated with this process are not accounted for in this plot.\gf{change plot }}
    \label{fig:alignment_plots_all}
\end{figure}

Figure~\ref{fig:alignment_plots_all} plots the average error rate across all of the datasets as a function of the floating point operations (FLOPS).

\section{Reward Model Training Configuration}
\label{appendix:rmtraining}
We trained a \citet{bt} reward model using \textsc{Llama-3.1-8B-Instruct} \citep{llama3} as the base model on a preference dataset for mathematical reasoning. The dataset contained 64 model-generated responses per prompt from GSM8K \citep{gsm8k} (on-policy data from \textsc{Llama-3.1-8B-Instruct}), from which we constructed one preference pair per prompt by selecting one ground truth correct answer as preferred and one incorrect answer as dispreferred. 
Training hyperparameters are detailed in Table~\ref{tab:hyperparameters}.

\begin{table}[h]
\centering
\caption{Training Hyperparameters for the RM for the task-specific experiments.}
\label{tab:hyperparameters}
\begin{tabular}{ll}
\toprule
\textbf{Parameter} & \textbf{Value} \\
\midrule
\multicolumn{2}{l}{\textbf{Model Configuration}} \\

Fine-tuning Type & Full \\
Chat Template & llama3 \\
\midrule
\multicolumn{2}{l}{\textbf{Dataset}} \\
Validation Split & 5\% \\
\midrule
\multicolumn{2}{l}{\textbf{Optimization}} \\
Learning Rate & $1 \times 10^{-5}$ \\
Total Batch Size & 128 \\
Training Epochs & 1.0 \\
LR Scheduler & Linear Decay\\
Warmup Ratio & 0.03 \\
Weight Decay & 0.0 \\
Precision & bfloat16 \\
\midrule
\multicolumn{2}{l}{\textbf{Infrastructure}} \\
Optimization Framework & DeepSpeed ZeRO Stage 3 \\
GPUS & 8 L40S\\
\bottomrule
\end{tabular}
\end{table}

\end{document}